%% file: savaal_acl.tex
\pdfoutput=1
\documentclass[11pt]{article}

\usepackage[final]{acl}
\usepackage{times}
\usepackage{latexsym}
\usepackage{xspace}
\usepackage{enumitem}
\usepackage{tabto}
\usepackage{subcaption}
\usepackage{tabularx}
\usepackage{array}
\usepackage{placeins}

\usepackage[T1]{fontenc}
\usepackage[utf8]{inputenc}
\usepackage{csquotes}

\usepackage{microtype}

\usepackage{inconsolata}

\usepackage{graphicx}
\usepackage{multirow}
\usepackage{amsmath}
\usepackage{amssymb}
\usepackage{mathtools}
\usepackage{amsthm}
\usepackage{caption} 
\usepackage{tcolorbox}
\usepackage{aliascnt}
\usepackage{xcolor}
\usepackage{tikz}
\usepackage{colortbl}
\usepackage{pgf} % Ensure this package is included in your preamble
\usepackage{pgfmath}
\usepackage{ragged2e}

\makeatletter
\renewcommand{\sectionautorefname}{\S\@gobble}
\makeatother % Changes 'subs
\makeatletter
\renewcommand{\subsectionautorefname}{\S\@gobble}
\makeatother % Changes 'subs
\makeatletter
\renewcommand{\subsubsectionautorefname}{\S\@gobble} % Changes 'subs
\makeatother
\def\compactify{\itemsep=0pt \topsep=0pt \partopsep=0pt \parsep=0pt}
\let\latexusecounter=\usecounter
\newenvironment{CompactEnumerate}
  {\def\usecounter{\compactify\latexusecounter}
   \begin{enumerate}}
  {\end{enumerate}\let\usecounter=\latexusecounter}

\tcbset{
    colframe=gray!40, 
    colback=gray!5,   
    coltitle=black,   
    fonttitle=\bfseries,
    sharp corners,
    boxrule=0.5mm,  
    width=\columnwidth,
    left=0.1mm,        
    right=0.1mm,      
    toptitle=0.1mm,
    bottomtitle=0.1mm,
    title=Question Generation Prompt
}
\title{\name: Scalable Concept-Driven Question Generation\\to Enhance Human Learning}

\author{
 \textbf{Kimia Noorbakhsh\textsuperscript{*}}, 
 \textbf{Joseph Chandler\textsuperscript{*}}, 
 \textbf{Pantea Karimi\textsuperscript{*}}, 
 \\
 \textbf{Mohammad Alizadeh}, 
 \textbf{Hari Balakrishnan}
\\
\\
 M.I.T. Computer Science and Artificial Intelligence Lab (CSAIL)
}

\definecolor{customblue}{HTML}{DAE8FC}
\definecolor{customred}{HTML}{F8CECC}
\definecolor{customgreen}{HTML}{D5E8D4}
\definecolor{custompurple}{HTML}{E1D5E7}
\definecolor{customorange}{HTML}{FFE6CC}

\newcommand*\circled[2]{\tikz[baseline=(char.base)]{
            \node[shape=circle, line width=0.75pt, draw=black, fill=#2, inner sep=1pt] 
            (char) {\textcolor{black}{\small\textsf{#1}}};}}

\newcommand{\bluecircle}{\raisebox{0pt}{\protect\tikz[baseline=(char.base)]{\protect\node[shape=circle,draw, fill=customblue, minimum size=1.8mm, inner sep=1pt] (char) {\footnotesize 1};}}}
\newcommand{\redcircle}{\raisebox{0pt}{\protect\tikz[baseline=(char.base)]{\protect\node[shape=circle,draw, fill=customred, minimum size=1.8mm, inner sep=1pt] (char) {\footnotesize 2};}}}
\newcommand{\greencircle}{\raisebox{0pt}{\protect\tikz[baseline=(char.base)]{\protect\node[shape=circle,draw, fill=customgreen, minimum size=1.8mm, inner sep=1pt] (char) {\footnotesize 3};}}}
\newcommand{\purplecircle}{\raisebox{0pt}{\protect\tikz[baseline=(char.base)]{\protect\node[shape=circle,draw, fill=custompurple, minimum size=1.8mm, inner sep=1pt] (char) {\footnotesize 4};}}}
\newcommand{\orangecircle}{\raisebox{0pt}{\protect\tikz[baseline=(char.base)]{\protect\node[shape=circle,draw, fill=customorange, minimum size=1.8mm, inner sep=1pt] (char) {\footnotesize 5};}}}

\newcommand*\sfboxed[1]{\tikz[baseline=(char.base)]{
            \node[shape=rectangle,line width=0.75pt, draw=black,inner sep=2pt, rounded corners=2pt] (char) {\textcolor{black}{\small\textsf{#1}}};}}
\newcommand{\NewPara}[1]{\noindent{\bf #1}}

\newcommand{\TheSystem}{Savaal\xspace}
\newcommand{\Baseline}{Direct\xspace}
\newcommand{\BaselineMT}{Direct\xspace}
\newcommand{\name}{Savaal\xspace}
\newcommand{\arxiv}{arXiv\xspace}
\newcommand{\gpt}{GPT-4o}

\begin{document}
\maketitle
\def\thefootnote{*}\footnotetext{These authors contributed equally to this work.}\def\thefootnote{\arabic{footnote}}
\input{010values}
\begin{abstract}
\input{000abstract}
\end{abstract}

\input{010newintro}

\input{020insights}
\input{030pipeline}
\input{040evaluation}
\input{050related_work}
\input{060conclusions}
\input{limitations}

\section*{Acknowledgments}
We thank all the expert evaluators for their time, insights, and feedback. This work was funded in part by Quanta Computer, Inc. under the AIR Project.

\newpage
% \bibliography{refs}

\input{savaal_acl.bbl}
\input{070appendix}

\end{document}

%% file: 010values.tex
% Scatter plot PhD

\newcommand{\hitresponse}{38\%\xspace}

\newcommand{\understandingpreferration}{6.5$\times$\xspace}
\newcommand{\choicespreferration}{3$\times$\xspace}
\newcommand{\usabilitypreferration}{2$\times$\xspace}

\newcommand{\savaalunderstandingpreferrationpercent}{61.9\%\xspace}
\newcommand{\baselineunderstandingpreferrationpercent}{9.5\%\xspace}
\newcommand{\sameunderstandingpreferrationpercent}{28.6\%\xspace}

\newcommand{\savaalchoicespreferrationpercent}{57.1\%\xspace}
\newcommand{\baselinechoicespreferrationpercent}{19.0\%\xspace}
\newcommand{\samechoicespreferrationpercent}{23.9\%\xspace}

\newcommand{\savaalusabilitypreferrationpercent}{47.6\%\xspace}
\newcommand{\baselineusabilitypreferrationpercent}{23.8\%\xspace}
\newcommand{\sameusabilitypreferrationpercent}{28.6\%\xspace}

% Weighted average difference
\newcommand{\phdunderstandingWAD}{17\%\xspace}
\newcommand{\phdchoicesWAD}{10\%\xspace}
\newcommand{\phdusabilityWAD}{11.4\%\xspace}

% PhD

\newcommand{\baselinephdunderstandingvalue}{29.0}
\newcommand{\savaalphdunderstandingvalue}{11.9}

\newcommand{\baselinephdchoicesvalue}{39.0}
\newcommand{\savaalphdchoicesvalue}{29.0}

\newcommand{\baselinephdusabilityvalue}{32.9}
\newcommand{\savaalphdusabilityvalue}{21.4}

% Paper 
\newcommand{\baselinepaperunderstandingvalue}{16.7}
\newcommand{\savaalpaperunderstandingvalue}{10.9}

\newcommand{\baselinepaperchoicesvalue}{22.1}
\newcommand{\savaalpaperchoicesvalue}{21.8}

\newcommand{\baselinepaperusabilityvalue}{15.3}
\newcommand{\savaalpaperusabilityvalue}{13.8}

%%%%%%%%%%% CHNAGEABLES UP %%%%%%%%%%%

% Percent disagree macros
\newcommand{\savaalphdusability}{\savaalphdusabilityvalue\%\xspace}
\newcommand{\baselinephdusability}{\baselinephdusabilityvalue\%\xspace}
\newcommand{\phdusabilityreductionvalue}{%
  \pgfmathparse{round(\baselinephdusabilityvalue/\savaalphdusabilityvalue*100 )/100}%
  \pgfmathprintnumber[fixed, precision=1]{\pgfmathresult}%
}

\newcommand{\phdusabilityreduction}{\phdusabilityreductionvalue$\times$\xspace}

\newcommand{\savaalphdchoice}{\savaalphdchoicevalue\%\xspace}
\newcommand{\baselinephdchoice}{\baselinephdchoicevalue\%\xspace}
\newcommand{\phdchoicereductionvalue}{%
  \pgfmathparse{round(\baselinephdchoicevalue/\savaalphdchoicevalue*100 )/100}%
  \pgfmathprintnumber[fixed, precision=1]{\pgfmathresult}%
}

\newcommand{\phdchoicereduction}{\phdchoicereductionvalue$\times$\xspace}

\newcommand{\savaalphdunderstanding}{\savaalphdunderstandingvalue\%\xspace}
\newcommand{\baselinephdunderstanding}{\baselinephdunderstandingvalue\%\xspace}
\newcommand{\phdunderstandingreductionvalue}{%
  \pgfmathparse{round(\baselinephdunderstandingvalue/\savaalphdunderstandingvalue*100 )/100}%
  \pgfmathprintnumber[fixed, precision=1]{\pgfmathresult}%
}
\newcommand{\phdunderstandingreduction}{\phdunderstandingreductionvalue$\times$\xspace}

\newcommand{\savaalphdchoices}{\savaalphdchoicesvalue\%\xspace}
\newcommand{\baselinephdchoices}{\baselinephdchoicesvalue\%\xspace}
\newcommand{\phdchoicesreductionvalue}{%
  \pgfmathparse{round(\baselinephdchoicesvalue/\savaalphdchoicesvalue*100 )/100}%
  \pgfmathprintnumber[fixed, precision=1]{\pgfmathresult}%
}
\newcommand{\phdchoicesreduction}{\phdchoicesreductionvalue$\times$\xspace}

\newcommand{\savaalpaperusability}{\savaalpaperusabilityvalue\%\xspace}
\newcommand{\baselinepaperusability}{\baselinepaperusabilityvalue\%\xspace}
\newcommand{\paperusabilityreductionvalue}{%
  \pgfmathparse{round(\baselinepaperusabilityvalue/\savaalpaperusabilityvalue*100 )/100}%
  \pgfmathprintnumber[fixed, precision=1]{\pgfmathresult}%
}
\newcommand{\paperusabilityreduction}{\paperusabilityreductionvalue$\times$\xspace}

\newcommand{\savaalpaperunderstanding}{\savaalpaperunderstandingvalue\%\xspace}
\newcommand{\baselinepaperunderstanding}{\baselinepaperunderstandingvalue\%\xspace}
\newcommand{\paperunderstandingreductionvalue}{%
  \pgfmathparse{round(\baselinepaperunderstandingvalue/\savaalpaperunderstandingvalue*100 )/100}%
  \pgfmathprintnumber[fixed, precision=1]{\pgfmathresult}%
}

\newcommand{\paperunderstandingreduction}{\paperunderstandingreductionvalue$\times$\xspace}

\newcommand{\savaalpaperchoices}{\savaalpaperchoicesvalue\%\xspace}
\newcommand{\baselinepaperchoices}{\baselinepaperchoicesvalue\%\xspace}
\newcommand{\paperchoicesreductionvalue}{%
  \pgfmathparse{round(\baselinepaperchoicesvalue/\savaalpaperchoicesvalue*100 )/100}%
  \pgfmathprintnumber[fixed, precision=1]{\pgfmathresult}%
}

\newcommand{\paperchoicesreduction}{\paperchoicesreductionvalue$\times$\xspace}

% Num evaluators macros
\newcommand{\numemails}{200\xspace}

\newcommand{\numpaperevaluatorsvalue}{55}
\newcommand{\numphdevaluatorsvalue}{21}

\newcommand{\numpapersvalue}{50}
\newcommand{\numphdvalue}{21}

\newcommand{\numloogle}{48\xspace}
\newcommand{\numfields}{5\xspace}

% Scale of dataset macros
\newcommand{\avgphdpagesvalue}{142}
\newcommand{\avgphdpages}{\avgphdpagesvalue\xspace}
\newcommand{\avgpaperpagesvalue}{19}
\newcommand{\avgpaperpages}{\avgpaperpagesvalue\xspace}

% Math relations
\newcommand{\numphd}{\numphdvalue\xspace}
\newcommand{\numpapers}{\numpapersvalue\xspace}

\newcommand{\totaldocuments}{\pgfmathparse{int(\numphdvalue+\numpapersvalue)}\pgfmathresult\xspace}%

\newcommand{\numpaperevaluators}{\numpaperevaluatorsvalue\xspace}
\newcommand{\numphdevaluators}{\numphdevaluatorsvalue\xspace}

\newcommand{\totalevaluatorsvalue}{\pgfmathparse{int(\numpaperevaluatorsvalue+\numphdevaluatorsvalue)}\pgfmathresult}

\newcommand{\numpaperquestionsvalue}{\pgfmathparse{int(\numpaperevaluatorsvalue*20)}\pgfmathresult}

\newcommand{\numphdquestionsvalue}{\pgfmathparse{int(\numphdevaluatorsvalue*20)}\pgfmathresult}

\newcommand{\totalevaluators}{\totalevaluatorsvalue\xspace}
\newcommand{\numpaperquestions}{\numpaperquestionsvalue\xspace}
\newcommand{\numphdquestions}{\numphdquestionsvalue\xspace}

\newcommand{\numtotalhumanquestionvalue}{\pgfmathparse{int(\numphdevaluatorsvalue*20 + \numpaperevaluatorsvalue*20)}\pgfmathresult}

\newcommand{\numtotalhumanquestion}
{\numtotalhumanquestionvalue\xspace}

\newcommand{\savaalcostreduction}{1.39$\times$\xspace}
\newcommand{\directcostinflation}{1.64$\times$\xspace}

%% file: 000abstract.tex
Assessing and enhancing human learning through question-answering is vital, yet automating this process remains challenging. While large language models (LLMs) excel at summarization and query responses, their ability to generate meaningful questions for learners is underexplored.

We propose \name,\footnote{\name means ``question'' in Hindi, and has a similar root in Persian and Arabic.} a scalable question-generation system with three objectives: (i) {\em scalability}, enabling question-generation from hundreds of pages of text (ii) {\em depth of understanding}, producing questions beyond factual recall to test conceptual reasoning, and (iii) {\em domain-independence}, automatically generating questions across diverse knowledge areas. Instead of providing an LLM with large documents as context, \name improves results with a three-stage processing pipeline. 
Our evaluation with \totalevaluators human experts on \totaldocuments papers and PhD dissertations shows that \name generates questions that better test depth of understanding by 6.5$\times$ for dissertations and 1.5$\times$ for papers compared to a direct-prompting LLM baseline. Notably, as document length increases, \name's advantages in higher question quality and lower cost become more pronounced.

%% file: 010newintro.tex
\section{Introduction}
\label{sec:intro}

Many people learn new material effectively by taking quizzes. Answering questions not only assesses knowledge, but also  reinforces learning by strengthening correct responses and revealing gaps in understanding. A major challenge in the 21st century is the rapid expansion of knowledge across fields like science, technology, medicine, law, finance, and more. AI tools are accelerating this growth, making it increasingly difficult for students, researchers, and professionals---from engineers to salespeople---to stay current. The need to learn efficiently and at scale has never been greater.

One response is to rely on AI for answers, effectively outsourcing expertise. While sometimes necessary, this does little to improve human understanding. Instead, we advocate using AI to enhance {\em our} ability to learn and master new material. %, and our work aims to advance this goal.

Programs like ChatGPT, Gemini, Claude, NotebookLM, Perplexity, and DeepSeek built atop large language models (LLMs) have made remarkable strides in summarization and question-answering. However, less attention has been given to {\em question generation}, specifically, creating high-quality questions that test human understanding and mastery of knowledge. That is the focus of this paper.

Anyone who has made an exam knows how difficult and time-consuming it is to make a good set of questions. Our goal is to produce questions automatically with three objectives:
\begin{CompactEnumerate}
\item {\em Scalability}: Generating questions across vast document corpora, such as rapidly evolving research fields or enterprise knowledge bases.
\item {\em Depth of understanding}: Producing questions beyond memorization and the superficial, requiring conceptual reasoning, synthesis, and analysis.
\item {\em Domain-independence}: Creating high-quality questions across diverse fields, including new material absent in an LLM’s pre-training data.
\end{CompactEnumerate}

Prior work on question generation has produced a small number of questions from short passages, but has not demonstrated scalability~\citep{du-etal-2017-learning, Neural_QG, chan-fan-2019-bert, li-etal-2021-addressing-semantic, knowledge_base_prompting, reading_comprehension_language_llm, code_QG, mcq_mult_sentence}. Our results (\autoref{sec:eval}) show that even well-engineered prompts to an LLM produce poor, repetitive questions on large text contexts (tens to hundreds of pages), highlighting the scalability challenge.

We present \textbf{\name}, a scalable question generation system for large documents. Savaal uses a three-stage pipeline. The first stage extracts and ranks the key concepts in a corpus of documents\footnote{We use ``document'' to also refer to the corpus of documents used to generate a quiz.} using a map-reduce computation. The second stage retrieves relevant passages corresponding to each concept with an efficient vector embedding retrieval model such as ColBERT~\cite{colbert}. Finally, the third stage prompts an LLM to generate questions for each ranked concept using the retrieved passages as context.

This approach scales well because each LLM computation is confined to a distinct, self-contained task while operating within a manageable context size. By first identifying core concepts and later synthesizing questions from all relevant passages, \name ensures that the generated questions are both targeted and conceptually rich, requiring deeper understanding by linking a given concept across different sections of a document.

We compare \name to a direct-prompting baseline (\Baseline) using \totalevaluators human expert evaluators (the primary authors of \numpapers recent conference papers and \numphd PhD dissertations in subfields of computer science and aeronautics) on \numtotalhumanquestion questions. We also evaluate each paper, as well as 48 arXiv papers, using an LLM as an AI judge.
%and also using an LLM judge on \numloogle papers from the Loogle \cite{loogle} dataset and \arxiv. 
%These evaluations span \numfields distinct fields. 
We find that: 
\begin{CompactEnumerate}
    \item On \numphdquestions questions from \numphdevaluators large documents (dissertations with average \avgphdpages pages), experts reported that \baselinephdunderstanding of \Baseline's questions {\em did not} test understanding, compared to \savaalphdunderstanding of \name, a \phdunderstandingreduction improvement. 
    They reported that \baselinephdchoices of \Baseline's questions lacked good choice quality, compared to \name's \savaalphdchoices, improving by \phdchoicesreduction. They found \baselinephdusability of \Baseline's questions {\em unusable} in a quiz, compared to \savaalphdusability of \name's questions, a \phdusabilityreduction reduction. Moreover, among experts with a preference, \understandingpreferration more favored \name over baseline in understanding, \choicespreferration in choice quality, and \usabilitypreferration in usability.

    \item Even on shorter documents, experts rated \name better in terms of depth of understanding and usability. On \numpaperquestions questions from \numpapers conference papers, \numpaperevaluators experts reported that \baselinepaperunderstanding of baseline's questions {\em did not} test understanding, compared to \savaalpaperunderstanding of \name, a \paperunderstandingreduction improvement.

    \item \name is less expensive than \Baseline as the number of questions grows: \Baseline's cost for 100 questions generated from the dissertations is \directcostinflation higher than \name (\$0.47 vs. \$0.77 on average per document).

    \item There is a large gap between AI judgments and human evaluations. Despite several attempts to align the AI judge to human responses, scores remained misaligned.% with human expert evaluations.

\end{CompactEnumerate}

%% file: 020insights.tex
\section{Why is Generating Good Questions Hard?}
\label{sec:insights}

Our goal is to enhance human learning from large documents spanning dozens to hundreds of pages by generating multiple-choice questions. Multiple-choice questions are widely used in assessments, are easy to use by learners, and are easy to grade. The task involves generating a set of clear questions, each with four  choices and a correct answer.

High-quality questions assess {\em depth of understanding}, requiring conceptual reasoning and plausible choices (distractors) that challenge the learner. Beyond clarity and precision, our notion of a good question is one that could appear in an advanced quiz on the material as judged by a human expert. While this paper focuses on generating individual high-quality questions, effective quiz sessions should ensure {\em concept coverage} and {\em adapting the difficulty} to prior answers in the session, both avenues for future work.

The main challenge in scalable question generation using LLMs is selecting an appropriate context to use with LLM prompts. We examine four potential strategies: (i) using the full document corpus, (ii) dividing the corpus into sections, (iii) summarizing the corpus, and (iv) using content selection classifiers~\citep{context_Steuer, Context_diverse_hadifar}. Although each strategy has merits, we show that each strategy fails on at least one of our key objectives: {\em scalability}, {\em depth of understanding}, or {\em domain-independence}.

\newcommand{\questionbox}[1]{%
    \colorbox{customblue}{\parbox{0.97\linewidth}{\vspace{1pt}\textbf{#1}\vspace{1pt}}}
}

\begin{table*}[t]
\centering
\renewcommand{\arraystretch}{1.0} % Adjust row height for better readability
\begin{tabular}{|p{0.2\textwidth}|m{0.5\textwidth}|m{0.2\textwidth}|} 
\hline
\rowcolor{gray!20} % Light gray background for header row
\textbf{Context} & \textbf{Generated Question} & \textbf{Issue} \\ \hline
\footnotesize
\sfboxed{1} \textbf{Entire Document} & 
\footnotesize
\questionbox{What is the primary benefit of using the Adam optimizer in training the Transformer model?}
A. It reduces the need for dropout regularization. \newline
\textbf{B.} It adapts the learning rate based on the training step, improving convergence. \newline
C. It eliminates the need for positional encodings. \newline
D. It simplifies the model architecture by reducing the number of layers. 
&
\footnotesize
\textbf{Too general}: The question is about a basic property of the Adam optimizer rather than the key ideas of the paper.

$\Rightarrow$ Does not test depth of understanding
\\ \hline
\footnotesize
\sfboxed{2} \textbf{Document Section}& 
\footnotesize
\questionbox{In evaluating the performance and efficiency of the Transformer (big) model on the WMT 2014 English-to-French translation task, which of the following factors most significantly contributes to its ability to outperform previous models at a reduced training cost?}
A. The use of a dropout rate of 0.1 instead of 0.3, which enhances model regularization and reduces overfitting.\newline
B. The implementation of beam search with a beam size of 4 and a length penalty $\alpha$ = 0.6, which optimizes the translation output quality.\newline
\textbf{C.} The averaging of the last 20 checkpoints, which stabilizes the model's performance and improves translation accuracy.\newline
D. The reduction in training time to less than 1/4 of the previous state-of-the-art model, which directly correlates with improved BLEU scores.&
\footnotesize
\textbf{Irrelevant detail:} Because the method looks at one section at a time, it fixates on minutiae and irrelevant details (e.g., “averaging the last 20 checkpoints”) that may seem important in isolation, but are not.

$\Rightarrow$ Does not test depth of understanding
\\ \hline
\footnotesize
\sfboxed{3} \textbf{Document Summary} &
\footnotesize
\questionbox{How does the Transformer model address the challenge of learning dependencies between distant positions in sequences compared to models like ConvS2S and ByteNet?}
A. By using convolutional layers to capture long-range dependencies\newline
B. By increasing the number of layers in the encoder and decoder stacks\newline
C. By employing a recurrent neural network to process sequences\newline
\textbf{D.} By reducing the number of operations to a constant using self-attention mechanisms"
& 
\footnotesize
\textbf{Missing context:} The summary mentions ``...The Transformer model addresses this by reducing the number of operations to a constant, using self-attention mechanisms.'' which led the LLM design this incomplete question.

$\Rightarrow$ Leads to inaccurate questions
\\ \hline
\end{tabular}
\caption{Examples from the ``Attention Is All You Need'' paper \citep{attention_is_all_you_need} using three different context selection methods. The questions are drawn from three separate 20-question quizzes, each generated using a different method via OpenAI's API \citep{openai_api} with the \texttt{gpt-4o} model.}
\label{tab:bad-examples-attention}
\vspace{-10 pt}
\end{table*}

\subsection{Using the Entire Document Corpus}
\label{sec:insights-whole-context}

One approach is to provide the entire document as context to an LLM for quiz generation. However, this method has two major drawbacks.
First, as prior research shows~\citep{lost-in-the-middle}, LLMs allocate attention unevenly across long documents, focusing more on the beginning and end while largely neglecting the middle. 

Second, LLMs struggle to capture dependencies between different sections of a long document~\citep{loogle}, leading to superficial questions and missing key concepts. When we prompted OpenAI's \texttt{gpt-4o} model with the full text of the ``Attention Is All You Need'' paper~\citep{attention_is_all_you_need}, many of the 20 generated questions overlooked key ideas. See Example \sfboxed{1} in \autoref{tab:bad-examples-attention} for a question, which is not relevant to the paper's key ideas.

We found that LLMs struggle to follow instructions when the context length is large~\cite{gao2024insights}. For example, we instruct the LLM not to repeat questions. While it avoids repetition when generating a few questions, larger batches (e.g., 20 questions) often contain duplicates.

\subsection{Using Document Sections}
\label{sec:insights-section-context}

An alternative is to split the document into sections, generate a limited number of questions per section, and later combine them into a quiz. While this method mitigates long-context issues, it introduces {\em context fragmentation}: the LLM cannot connect concepts spanning multiple sections. It often misses deeper connections that can assess stronger conceptual understanding. For example, key insights in a paper’s Algorithm or Methods section may be essential for understanding its Results, but treating these sections independently leads to disjointed, narrow questions.

Another issue is {\em uneven importance weighting}. Not all sections contribute equally to the document’s  ideas, yet a naïve section-based approach may overemphasize minor details while missing key concepts. Example \sfboxed{2} in \autoref{tab:bad-examples-attention} shows how this can generate irrelevant memorization questions.

\subsection{Summarization}
\label{sec:insights-summary}

Providing a {\em document summary} as context offers another way to streamline question generation. While LLMs are effective at summarization, summaries often lack critical details, leading to vague or incomplete questions. More concerning, summaries can introduce hallucinations~\citep{llm_hallucination}, distorting or misrepresenting causal relationships and fabricating details, further degrading question quality.

Example \sfboxed{3} in \autoref{tab:bad-examples-attention} illustrates how summarization can result in misleading or imprecise questions. Here, the summary includes a statement about using self-attention to ``reduce the number of operations to a constant'', but omits that this refers to {\em sequential} operations and maximum path length (Sec. 4 of \citep{attention_is_all_you_need}), leading to an inaccurate question.

\subsection{Content Selection Classifiers}

Some methods attempt to select relevant content for question generation, often using trained models to identify key passages~\citep{context_Steuer, Context_diverse_hadifar}. However, these approaches typically require domain-specific training data (e.g., pre-existing question-answer pairs), making them {\em domain-dependent}. Such approaches are frequently limited in scope, making them neither reliable nor generalizable to diverse domains. 

%% file: 030pipeline.tex
\begin{figure*}[!t]
\centering
    \includegraphics[width=1\linewidth]{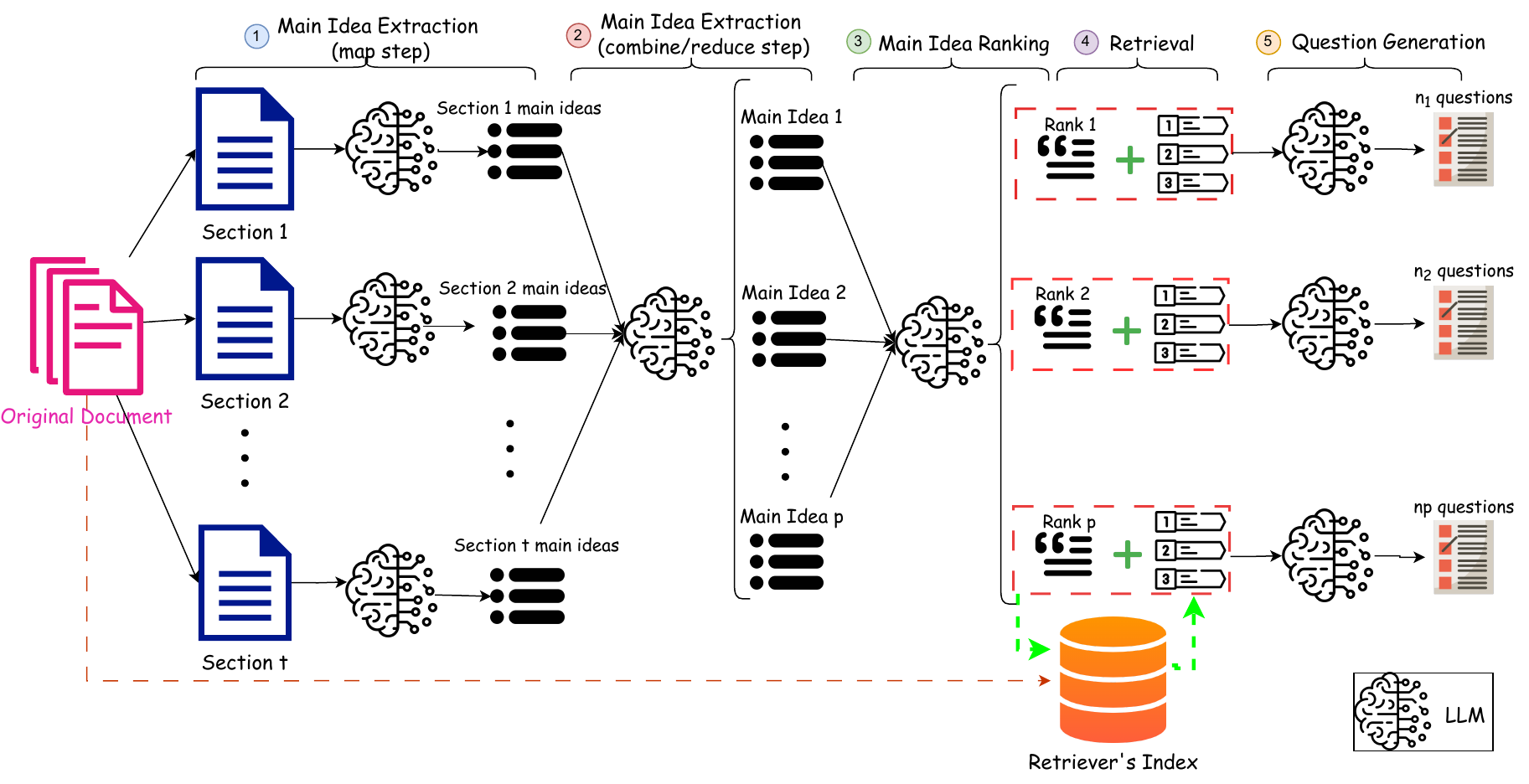}
    \caption{\name's Pipeline. \bluecircle\ \name extracts main ideas from sections of the document in parallel, \redcircle\ combines them into a succinct list, and \greencircle\ ranks them in order of importance. Next, \purplecircle\ \name fetches relevant passages from the document using a vector-based retrieval model. Finally, \orangecircle\ given a main idea and fetched passages, \name generates questions.}
    \label{fig:savaal-workflow}
\vspace{-10 pt}
\end{figure*}

\section{\name's Question-Generation Pipeline}
\label{sec:pipeline}

To address challenges of \autoref{sec:insights}, we propose a novel three-stage pipeline: \emph{main idea extraction}, \emph{relevant passage retrieval}, and \emph{question generation}. \autoref{fig:savaal-workflow} shows Savaal's workflow. The idea is to generate questions targeted at key explicitly determined concepts and to retrieve passages relevant to the concept from the source document.
% to generate questions.

\subsection{Extracting Main Ideas}
\label{sec:pipeline-main-idea}
This stage extracts succinct main ideas from different document chapters. This is done in a map-combine-reduce fashion~\cite{langchain_mapreduce}. First, we use GROBID~\citep{GROBID} to parse and segment documents into distinct sections.

In the map stage, \circled{1}{customblue}, we use an LLM to extract the main ideas for each section individually. These extracted main ideas are aggregated and deduplicated in the combine stage, \circled{2}{customred}, into a single, cohesive list of the paper’s main ideas. If the combined output exceeds a predefined length threshold (set to the maximum token window of the LLM), the reduce stage collapses the list further for brevity and clarity. The result is a curated list of main ideas, including main idea titles and their short descriptions (see \autoref{subsubsec:example_main_idea} for examples). The same (or a different) LLM then ranks the main ideas based on their importance in the ranking stage in \circled{3}{customgreen} (see \autoref{subsec:appendix_prompts} for the prompts).

Initially, we attempted to extract the main ideas for the entire document in one shot. However, as noted in \autoref{sec:insights-whole-context}, as the context length grew, this became less effective. We found that using map-reduce extracted main ideas that were more detailed and useful for question generation, particularly on large documents.

\subsection{Retrieving Relevant Passages}
\label{sec:pipeline-retrieval}

Because the main ideas in \autoref{sec:pipeline-main-idea} are concise, they lack sufficient content to generate a question. As discussed in \autoref{sec:insights-summary}, asking an LLM to generate questions based on a concept alone (a main idea or even a summary) has shortcomings. To overcome this problem, \name retrieves relevant text segments directly from the original document to provide granular content for generating a question and to ensure that the questions are grounded in truth.

\name's retriever uses ColBERT, a late-interaction retrieval method~\citep{colbert, colbertv2}, to find the most relevant passages for each main idea (stage \circled{4}{custompurple}).
% integrated in the RAGatouille \footnote{\url{https://github.com/AnswerDotAI/RAGatouille}} library.
For each ranked main idea in \circled{3}{customgreen}, we retrieve the top $k$ passages as added context for the next stage ($k=3$ in our experiments).

We chose ColBERT for its state-of-the-art performance and wide adoption, but any high-performing retrieval method could be used. We also tried using the LLM to identify passages related to a main idea, but as in \autoref{sec:insights-whole-context} and \autoref{sec:pipeline-main-idea}, it struggled with large context sizes.

\subsection{Generating Questions and Choices}
\label{sec:pipeline-QG}

After retrieving the passages for each main idea, stage \circled{5}{customorange} instructs an LLM to generate questions. To create $N$ questions from $M$ ideas, we generate $N/M$ questions per idea.\footnote{We use only the top $N$ ranked main ideas if $N < M$.}  The prompt (\autoref{fig:question_generation}) includes the main idea and its retrieved passages.

Although LLMs often produce good questions, generating good {\em choices} is more challenging. Poorly designed choices can make the correct answer too obvious or, worse, introduce ambiguity or multiple correct options. We experimented with many prompt variations to improve choice quality, yielding mixed results. We also tested a separate ``choice refinement'' stage, where an LLM was specifically instructed to improve the answer choices for a given question. This prompt included detailed constraints, such as ensuring alignment with the question's intent (e.g., a question about benefits should not include limitations as choices; see \autoref{appendix:choice-refine}).
Although this additional step produced more challenging choices, we found that it caused excessive ambiguity and was less preferred by human expert evaluators. Therefore, \name does not include a choice refinement stage. Instead, its question-generation prompt explicitly emphasizes that the choices should be ``plausible distractors''.

Finally, we observed {\em positional biases} in the placement of the correct choice, corroborating prior findings~\cite{pezeshkpour2023large}. For example, in a set of 1000 questions from 50 papers (20 per paper) generated by \texttt{GPT-4o}, choice B was correct 73.3\% of the time! Thus, we randomize the choices to eliminate this bias.

%% file: 040evaluation.tex
\section{Evaluation}
\label{sec:eval}

We evaluated \name on \totaldocuments documents using both human experts and an AI judge. We used \texttt{GPT-4o} via the OpenAI API as our primary LLM. We also evaluated \texttt{Meta-Llama-3.3-70B-Instruct} (\autoref{subsec:ablation-model}). All models are set to temperature 0.0 for all experiments, with default settings for all other parameters. \name is model-agnostic and is compatible with current LLMs. We implemented our pipeline using LangChain~\cite{langchain} and traced our experiments in Weights \& Biases~\cite{wandb}.
%any commercial or open-source LLM.

\subsection{Datasets}
\label{sec:evaluation-data}
%We gathered three datasets for evaluation:
\begin{itemize}[topsep=0pt, itemsep=0pt, leftmargin=*]
    \item \textbf{PhD dissertations}: \numphd long documents in Aerospace, Machine Learning, Networks, Systems, and Databases (\autoref{tab:human-eval-dataset-stats}).
    \item \textbf{Conference papers}: \numpapers papers from conferences in CS and Aeronautics in 2023 and 2024.
    \item \textbf{Diverse \arxiv papers}: \numloogle papers from CS, Physics, Mathematics, Economics, and Biology (\autoref{tab:benchmark-stats}). 
\end{itemize}

\begin{table}[h]
\centering
\renewcommand{\arraystretch}{1} % Increase row height for better readability
\setlength{\tabcolsep}{1pt} % Adjust column spacing for better fit
% \begin{small}
\begin{tabular}{|l|c|c|}
\hline
\small  \textbf{Statistic} & \small  \textbf{Conference Papers} & \small \textbf{Dissertations} \\
\hline
\small \textbf{No. Documents} & \small \numpapers & \small \numphd \\ 
\hline
\small \textbf{Avg. Words} & \small 10,354 & \small 26,511 \\ 
\hline
\small \textbf{Avg. Pages} & \small \avgpaperpages & \small \avgphdpages \\
\hline
\end{tabular}
\caption{Statistics for the number of words in the conference papers and PhD dissertations.}
\label{tab:human-eval-dataset-stats}
% \end{small}
\end{table}

\subsection{Methods Compared}
\label{sec:evaluation-baselines}
We compare \name to \Baseline, a direct-prompting baseline (\autoref{sec:insights-whole-context}) that provides the entire document to the LLM with a detailed prompt to generate $N$ multiple-choice questions (\autoref{fig:baseline_question_generation_prompt}). We found that when $N$ exceeds $\approx$ 20, \Baseline fails to produce $N$ distinct questions. Since broad concept coverage requires generating a large pool of questions and sampling for shorter quizzes, we generate $N > 20$ questions in batches of $b=20$ using an additional prompt (\autoref{fig:baseline_large_question_generation_prompt}). We use this {\em multi-turn method} for \Baseline on longer documents.

We evaluate other methods using the AI judge: Summary (\autoref{sec:insights-summary}) and Single-Prompt Savaal, which condenses Savaal's idea extraction, retrieval, and question generation into a single prompt (\autoref{subsec:ablation-methods}).

\subsection{Evaluation Criteria}
\label{sec:evaluation-metrics}

Evaluating the quality of questions is challenging because it involves subjective human judgment~\cite{fu2024qgeval}. We primarily rely on human evaluations but also use \texttt{GPT-4o} as an AI judge~\cite{naismith2023automated} to expand the scope of our evaluation to more approaches, documents, and criteria. 

% Exempt ID: E-6417

\paragraph{Human experts:} We generated 10 multiple-choice questions from Savaal and 10 from \Baseline for each of the \numphd PhD dissertations and \numpapers conference papers. We contacted the primary authors to evaluate the quality of questions via a secure web-based feedback form.\footnote{\emph{MIT} Institutional Review Board exempted this study (Exemption Number: E-6417). All the personnel were certified, and participants were over 18 years of age and provided informed consent before participating.} We asked each expert to rate their questions on clarity, depth of understanding\footnote{Used interchangeably with understanding.}, and quality of choices using a four-point scale: \emph{Disagree}, \emph{Somewhat Disagree}, \emph{Somewhat Agree}, and \emph{Agree}. They also assessed usability by answering: ``Would I use this question on a graduate-level quiz?'' with options: {\em Yes}, {\em Yes with small changes}, and {\em No}. The questions were randomly mixed and the evaluators were blind to their source. In all, \totalevaluators experts participated (\autoref{subsec:appendix_human_eval_conduct}).

\label{sec:metrics-auto}
\paragraph{AI judge:} We prompted \texttt{GPT-4o} at temperature 0.0 to score each question on a 1–4 scale (\autoref{subsubsec:eval-prompts}) on Depth of Understanding, Quality of Choices, Clarity, Usability, Difficulty, Cognitive Level, and Engagement (\autoref{subsec:ablation-metrics}). Our evaluation prompts provide detailed guidelines than those given to humans, including explicit criteria for each rating (\autoref{subsubsec:eval-prompts}).

\subsection{Results with Human Experts}
\label{sec:evaluation-results}

\begin{figure}[!t]
    \centering
    \begin{subfigure}[b]{0.9\linewidth}
        \centering
        \includegraphics[width=1\linewidth]{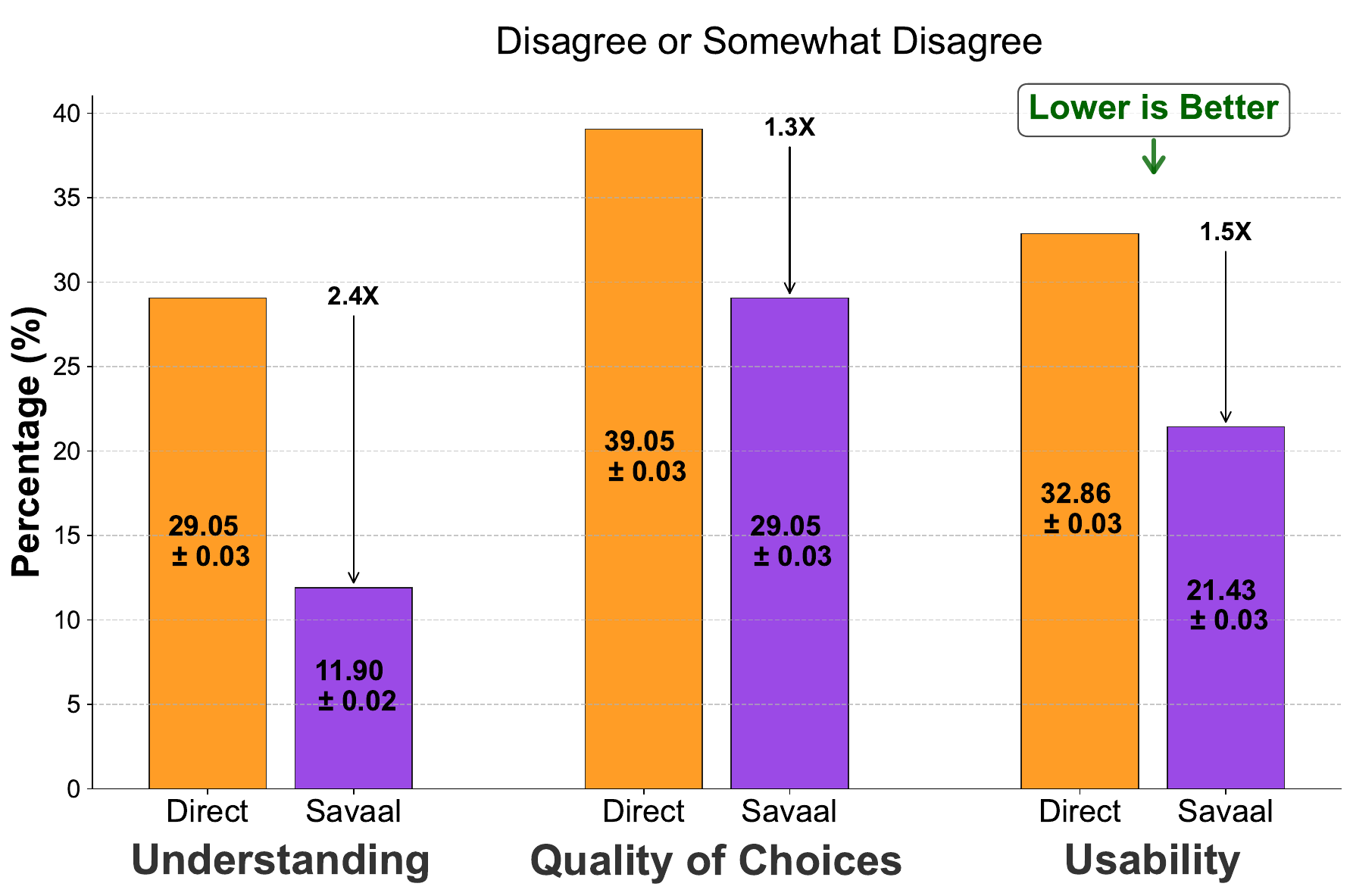}
        \caption{PhD dissertations: \numphdquestions questions, \numphdevaluators experts.}
        \label{fig:human-eval-disagree-phd}
    \end{subfigure}
    \hfill
    \begin{subfigure}[b]{0.9\linewidth}
        \centering
        \includegraphics[width=1\linewidth]{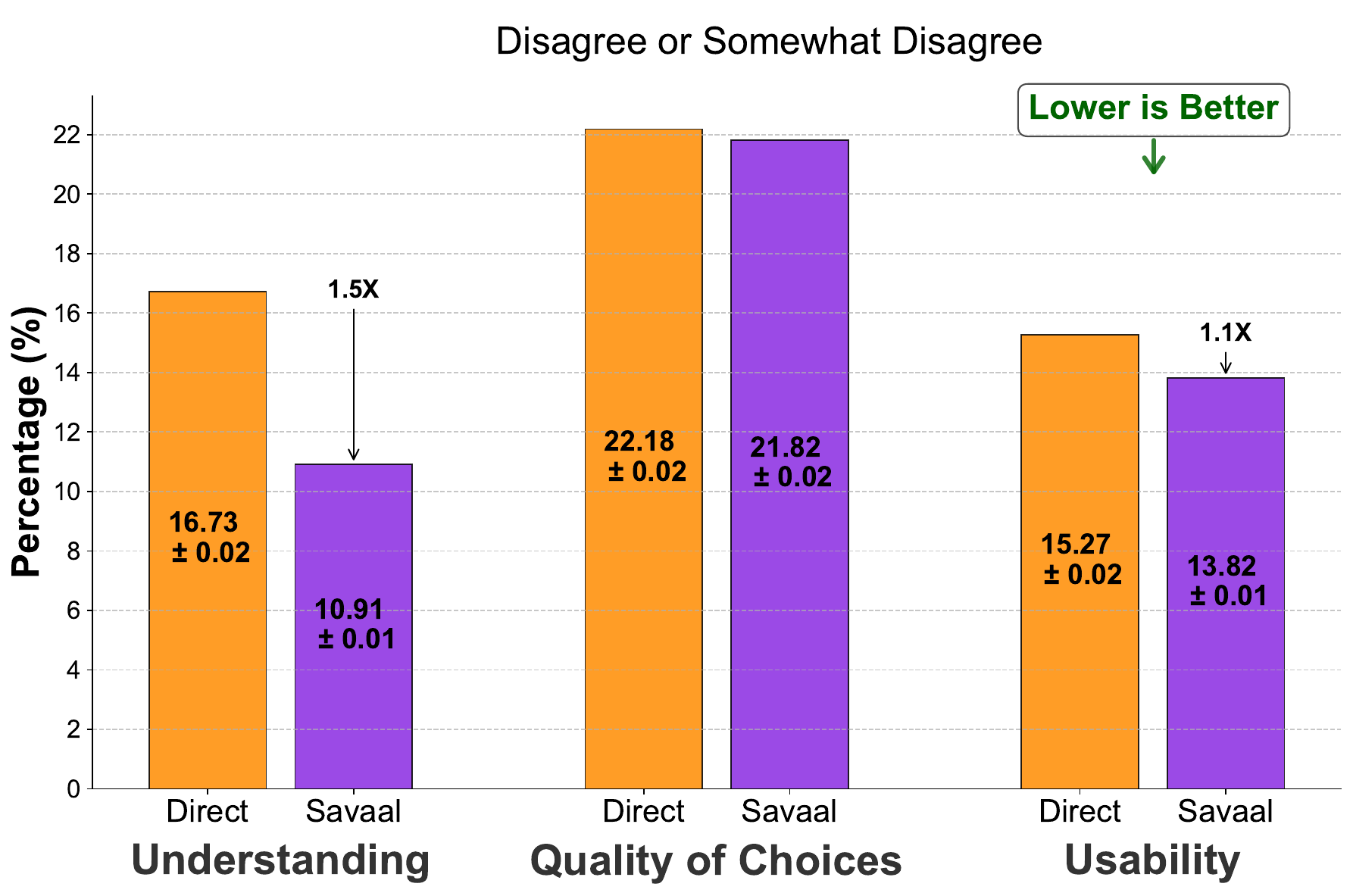}
        \caption{Conference papers: \numpaperquestions questions, \numpaperevaluators experts.}
        \label{fig:human-eval-disagree-paper}
    \end{subfigure}
    \vspace{-10 pt}
    \caption{ Summary of human evaluation: The charts show the percentage and standard error of respondents who {\em Disagree} or {\em Somewhat Disagree} with questions on understanding, choice quality, and usability. {\bf Lower values indicate better performance.}}
    \label{fig:human-eval-disagree}
\vspace{-20 pt}
\end{figure}

\begin{figure*}[!t]
    \centering
    \begin{subfigure}[b]{0.32\linewidth}
        \centering
        \includegraphics[width=\linewidth]{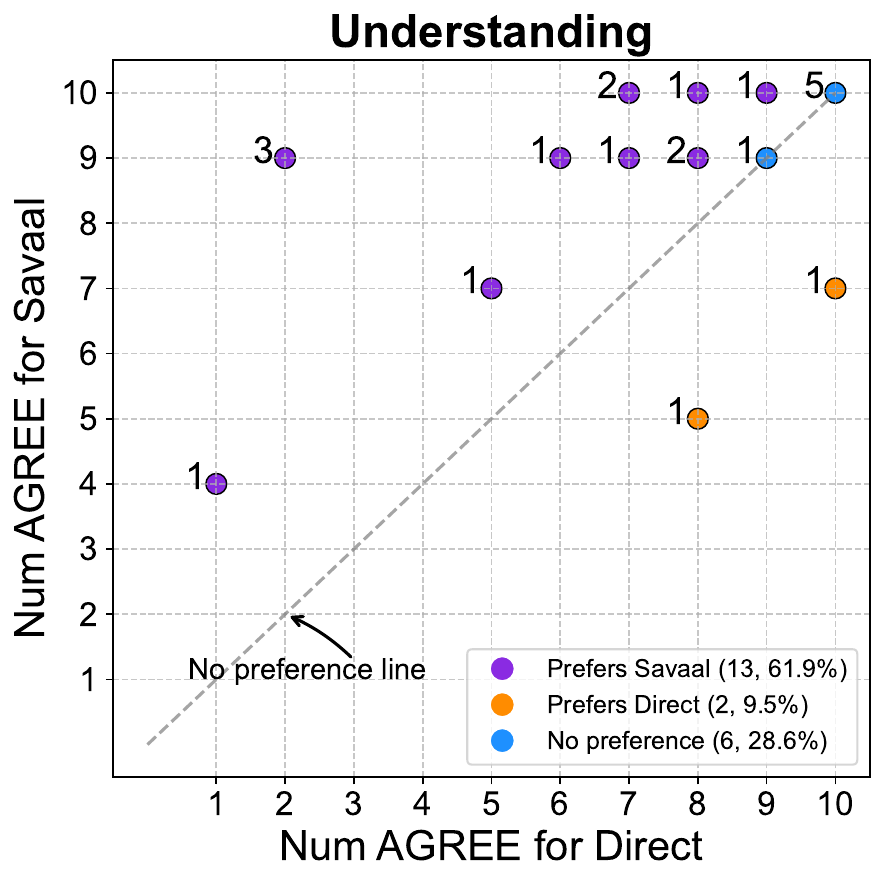}
        \caption{Depth of understanding: 61.9\% prefer \name, 9.5\% \Baseline.}
        \label{fig:thesis-scatter-understanding}
    \end{subfigure}
    \hfill
    \begin{subfigure}[b]{0.32\linewidth}
        \centering
        \includegraphics[width=\linewidth]{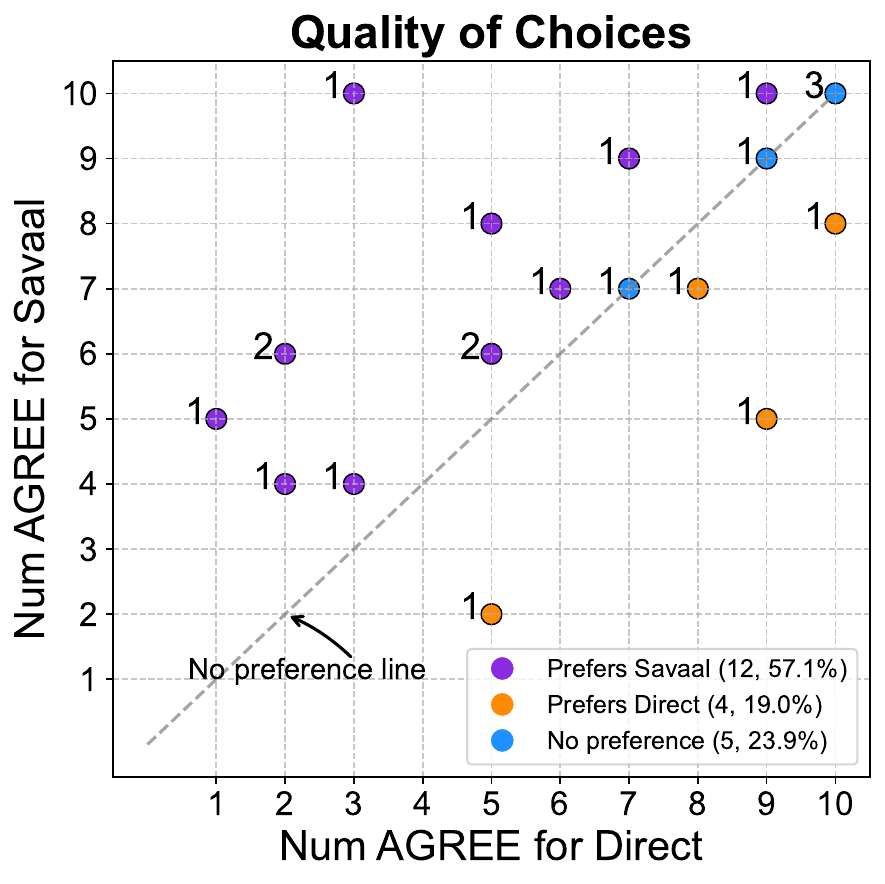}
        \caption{Quality of choices: 57.1\% prefer \name, 19\% \Baseline.}
        \label{fig:thesis-scatter-choices}
    \end{subfigure}
    \hfill
    \begin{subfigure}[b]{0.32\linewidth}
        \centering
        \includegraphics[width=\linewidth]{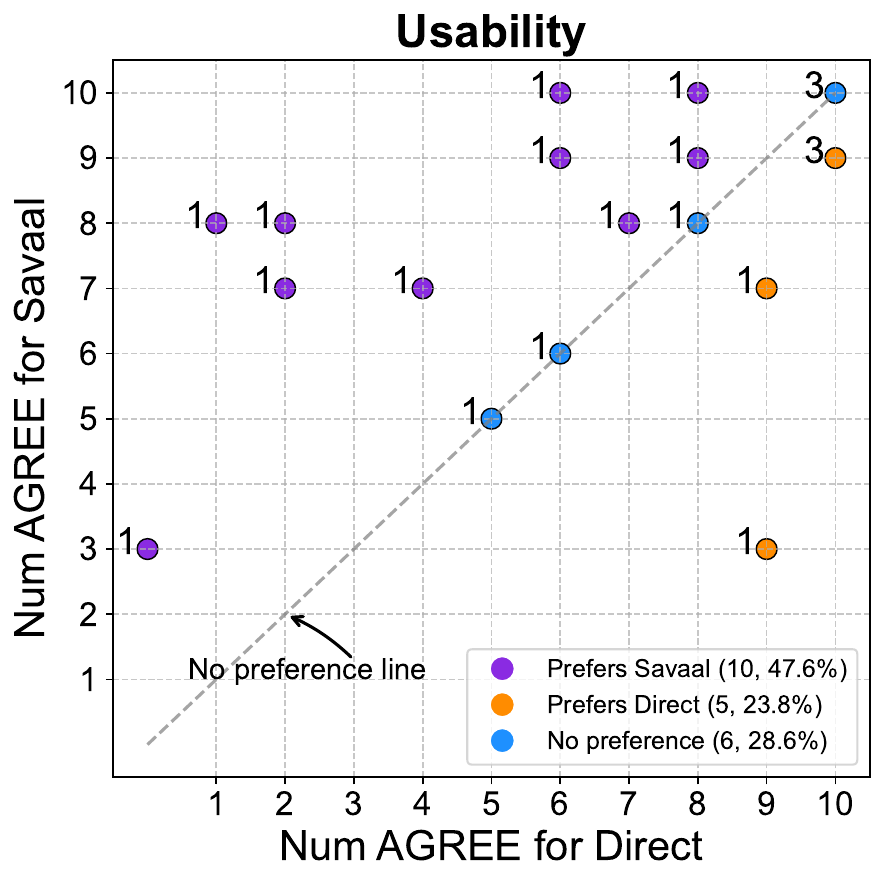}
        \caption{Usability: 47.6\% prefer \name, 23.8\% \Baseline.}
        \label{fig:thesis-scatter-overall}
    \end{subfigure}
    
    \caption{Expert preferences for \numphdevaluators PhD dissertations. Each point shows the number of \emph{Agree}s or \emph{Somewhat Agree}s in a 10-question quiz for each of \name and \Baseline. The majority of experts prefer \name to \Baseline on depth of understanding, quality of choices, and usability on long documents (experts above $y=x$ prefer \name).}
    \label{fig:human-eval-scatter}
    \vspace{-10 pt}
\end{figure*}

\label{sec:evaluation-human}

\autoref{fig:human-eval-disagree} summarizes the results of our expert human evaluation on PhD dissertations and papers. We show here the negative sentiment of the experts, i.e., the percentage of questions that experts responded with \emph{Disagree} or \emph{Somewhat Disagree} for each criterion (see \autoref{fig:human-phd-breakdown} and \autoref{fig:human_paper_breakdown} for the full breakdown). 

For the \numphdquestions questions from \numphd PhD dissertations (\autoref{fig:human-eval-disagree-phd}), the experts responded that \baselinephdunderstanding of \Baseline's questions {\em did not test understanding}; by contrast, only \savaalphdunderstanding of \name's questions did not, a  \phdunderstandingreduction reduction in negative sentiment. They also rated \baselinephdusability of \Baseline's questions as {\em unusable in a quiz}, versus \savaalphdusability for \name, a \phdusabilityreduction reduction.

For  conference papers (\autoref{fig:human-eval-disagree-paper}), on \numpaperquestions questions, \numpaperevaluators experts\footnote{Some papers had multiple expert respondents.} found that \savaalpaperunderstanding of \name's questions {\em did not} test understanding, versus \baselinepaperunderstanding for \Baseline, a \paperunderstandingreduction improvement. They also rated \baselinepaperusability of \Baseline's questions as {\em unusable}, versus \savaalpaperusability for \name.

The experts agreed or somewhat agreed that over 90\% of the questions in both \Baseline and \name had clarity (not shown in the figure). This result is unsurprising because LLMs can be prompted to generate coherent and unambiguous text. 

For PhD dissertations, \autoref{fig:human-eval-scatter} shows how each of the \numphdevaluators experts scored \name vs. \Baseline on the metrics for the PhD dissertations. The $x$ and $y$ axes show number of \emph{Agree} or \emph{Somewhat Agree} for \Baseline and \name, respectively. Each point represents one expert evaluator. 

We observe that \savaalunderstandingpreferrationpercent favor \name over \Baseline for understanding (\autoref{fig:thesis-scatter-understanding}), whereas only \baselineunderstandingpreferrationpercent (\understandingpreferration fewer) prefer \Baseline over \name (\sameunderstandingpreferrationpercent rate the two systems the same). For choice quality, \savaalchoicespreferrationpercent prefer \name compared to \baselinechoicespreferrationpercent for \Baseline (\choicespreferration more, see \autoref{fig:thesis-scatter-choices}), while for usability \savaalusabilitypreferrationpercent prefer \name compared to \baselineusabilitypreferrationpercent for \Baseline (\usabilitypreferration more, see \autoref{fig:thesis-scatter-overall}). 

The data in \autoref{fig:human-eval-scatter} also shows that, on average, expert evaluators rated \emph{Agree} or \emph{Somewhat Agree} for more questions in \name quizzes than \Baseline: \phdunderstandingWAD more for understanding, \phdchoicesWAD more for quality of choices, and \phdusabilityWAD more for usability.

\autoref{fig:human_paper_breakdown} shows the breakdown of expert responses for \numpaperquestions questions from the conference papers. On these shorter documents, experts slightly prefer \name over \Baseline in terms of depth of understanding. They reported that 16.7\% of \TheSystem's questions {\em did not} test understanding, compared to 10.9\% for \Baseline. Experts rated the two methods similarly for choice quality and usability. As in the results for Ph.D. dissertations (\autoref{fig:human-auto-correlation}), the \texttt{GPT-4o} scores (\autoref{fig:AI_paper_breakdown}) correlated poorly with expert evaluations.

\begin{figure*}[h]
    \centering
    \begin{subfigure}[b]{0.32\linewidth}
        \centering
        \includegraphics[width=\linewidth]{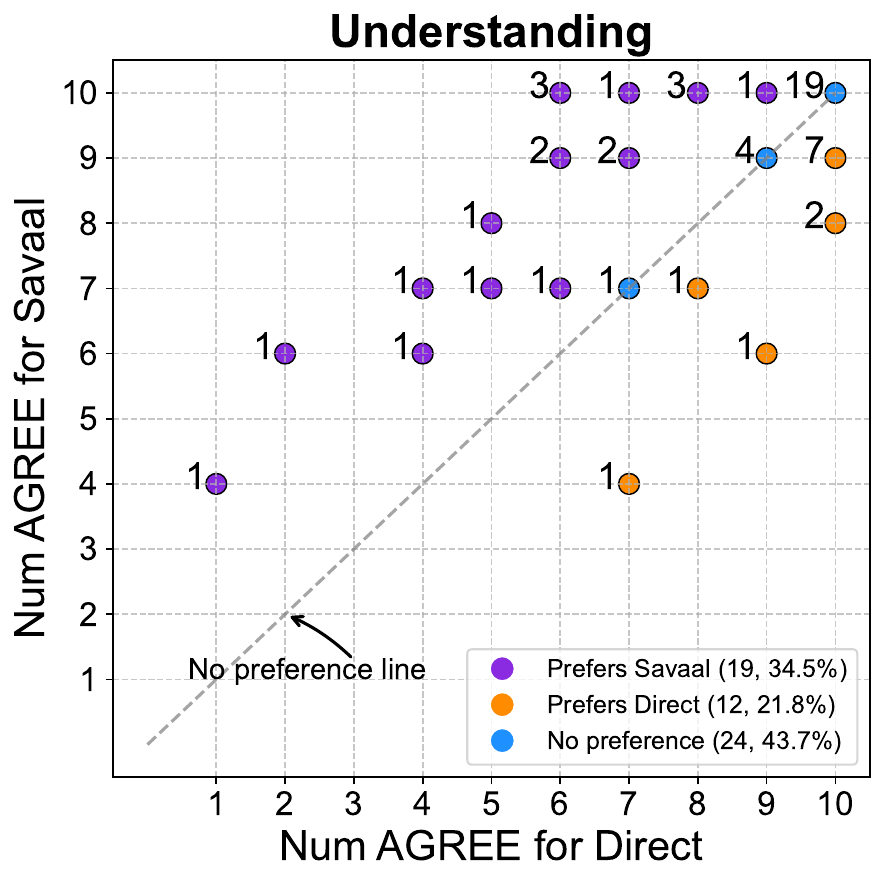}
        \caption{Depth of understanding: 34.5\% prefer \name, 21.8\% prefer \Baseline.}
        \label{fig:paper-scatter-understanding}
    \end{subfigure}
    \hfill
    \begin{subfigure}[b]{0.32\linewidth}
        \centering
        \includegraphics[width=\linewidth]{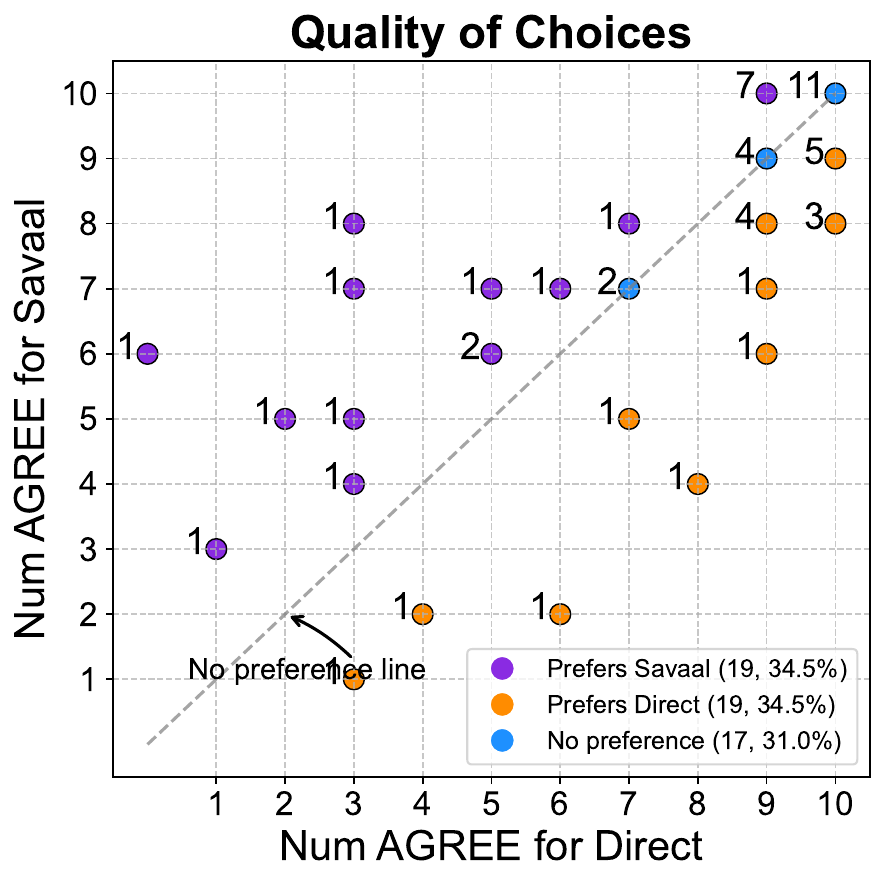}
        \caption{Quality of choices: no specific preference exhibited.}
        \label{fig:paper-scatter-choices}
    \end{subfigure}
    \hfill
    \begin{subfigure}[b]{0.32\linewidth}
        \centering
        \includegraphics[width=\linewidth]{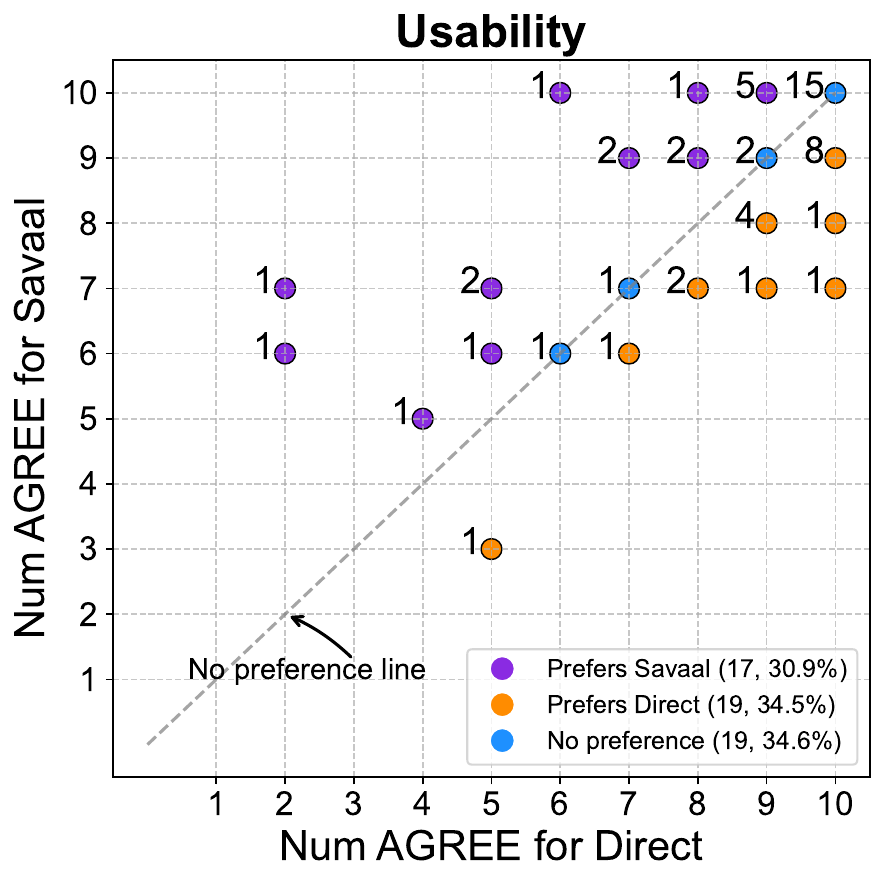}
        \caption{Usability: no specific preference exhibited.}
        \label{fig:paper-scatter-overall}
    \end{subfigure}
    
    \caption{Human expert preferences for \numpaperevaluators experts on short conference papers. Each point shows the number of \emph{Agree}s in a 10-question quiz for \name and \Baseline respectively. More experts prefer \name to \Baseline on the depth of understanding. Experts don't exhibit any preference between the quality of choices and usability on short documents (experts above $y=x$ prefer \name).}
    \label{fig:paper-human-eval-scatter}
\end{figure*}

\autoref{fig:paper-human-eval-scatter} shows how each of the \numpaperevaluators experts scored \name vs. \Baseline. The $x$-axis shows the number of \emph{Agree} or \emph{Somewhat Agree} for \Baseline, and the $y$-axis shows the same for \name. Each point represents one expert evaluator. Among evaluators with a preference, 1.5$\times$ more experts favor \TheSystem over \Baseline in understanding (34.5\% for \name vs 21.8\% for \Baseline, \autoref{fig:paper-scatter-understanding}). Experts do not exhibit a strong preference between \name and \Baseline for choice quality (\autoref{fig:paper-scatter-choices}) or usability (\autoref{fig:paper-scatter-overall}). The average relative increase in the Agree score for \TheSystem compared to \Baseline is 5.8\% for understanding, 4\% for quality of choices, and 1.5\% for usability.
% , meaning that on average, experts like at least one more question in \name's quizzes compared to \Baseline.

\subsection{Results with an AI Judge}
\label{sec:evaluation-auto}

We used an AI judge to scale evaluations across more documents and criteria. We first examined its alignment with human experts by having \texttt{GPT-4o} evaluate the same \numphdquestions questions from the expert-reviewed dissertations dataset. 

\begin{figure}[t]
    \centering
    \begin{subfigure}[b]{1\linewidth}
        \centering
        \includegraphics[width=1\linewidth]{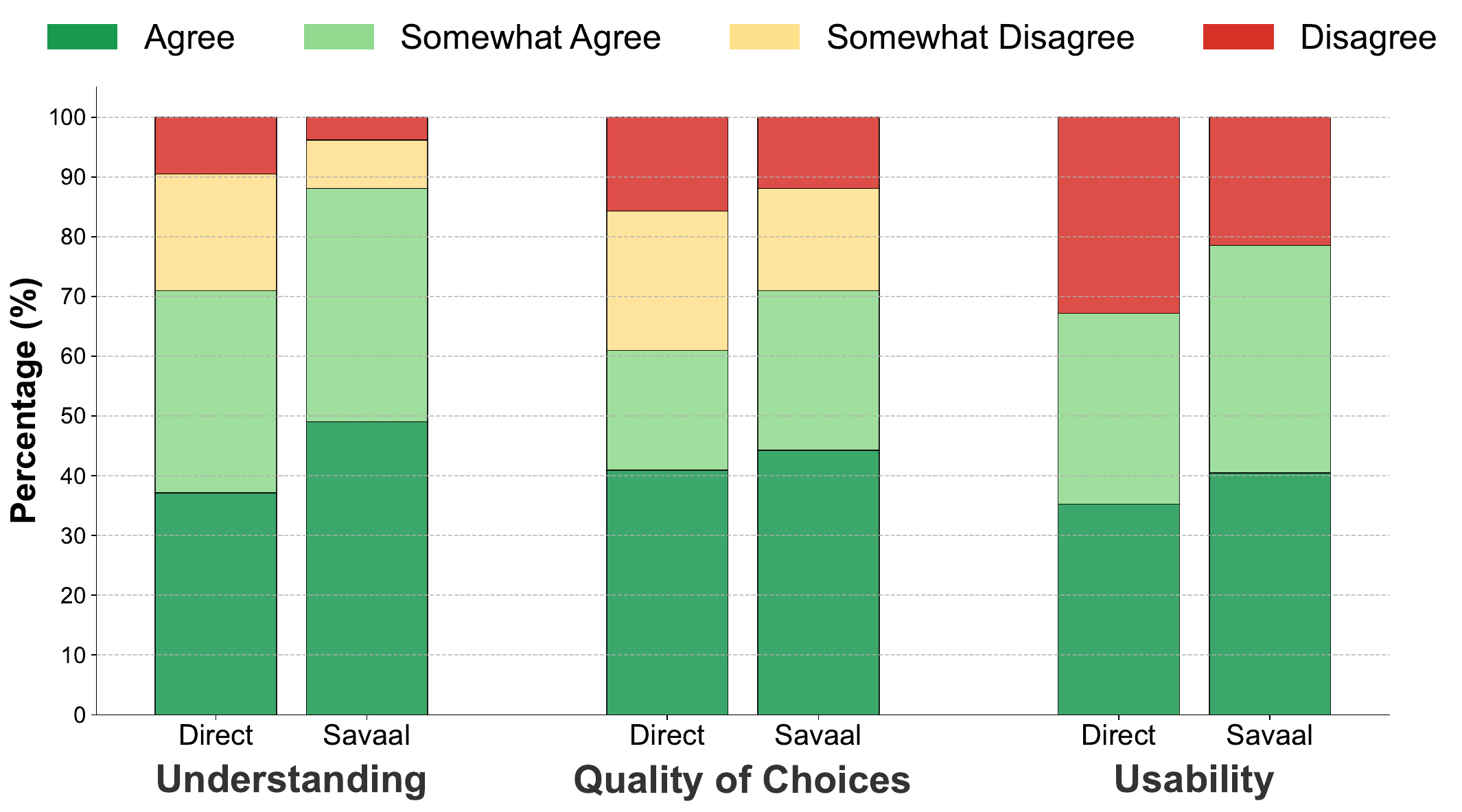}
        \caption{Breakdown of human expert scores on PhD dissertations.}
        \label{fig:human-phd-breakdown}
    \end{subfigure}  
    \hfill
    \begin{subfigure}[b]{1\linewidth}
        \centering
        \includegraphics[width=1\linewidth]{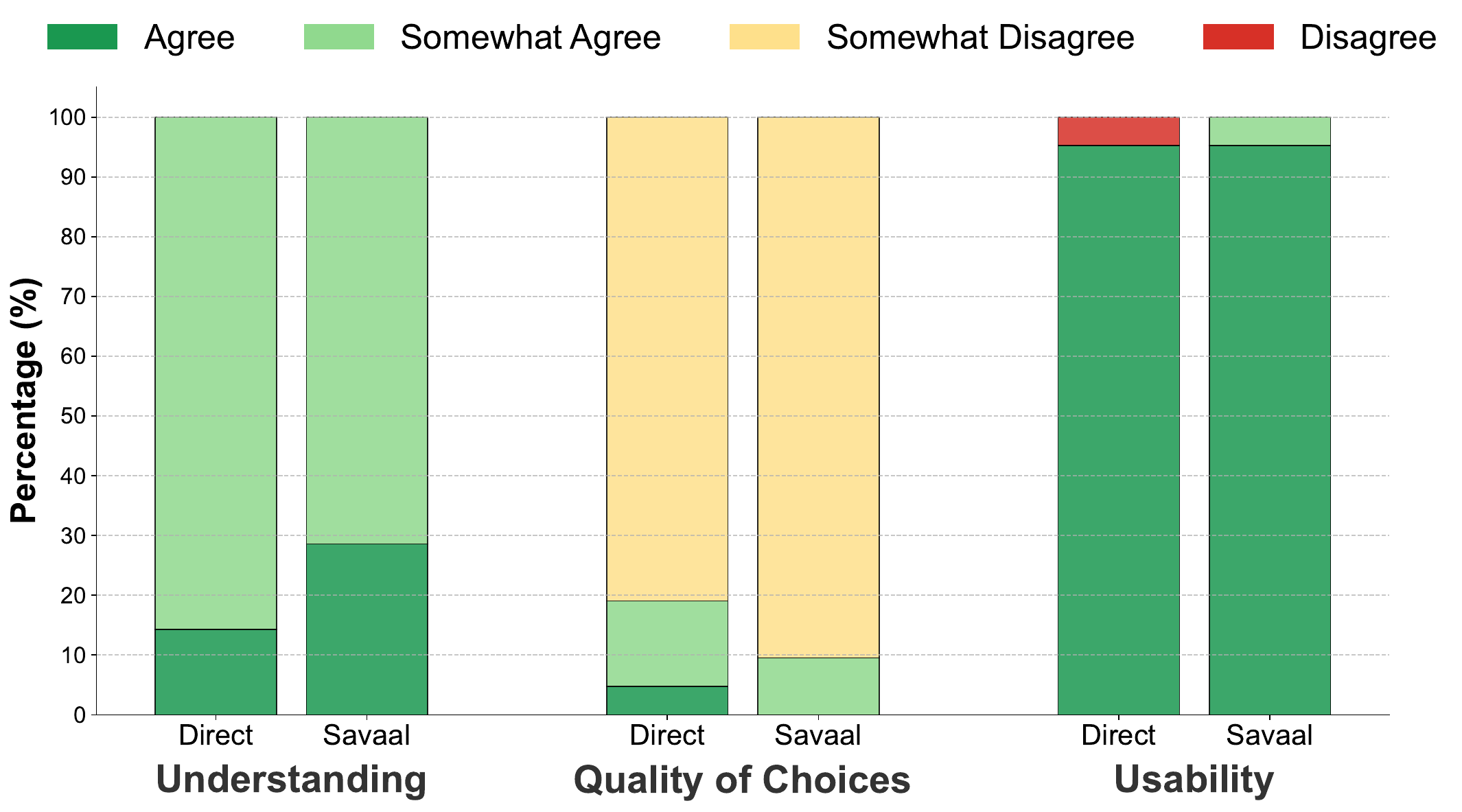}
        \caption{Breakdown of GPT-4o AI judge scores on PhD dissertations.}
        \label{fig:auto-correlation-ai}
    \end{subfigure}
    \caption{Score distribution for  \protect\numphdquestions questions from dissertations: GPT-4o as a judge does not align with humans for assessing the metrics.}
    \label{fig:human-auto-correlation}
    \vspace{-20 pt}
\end{figure}

\autoref{fig:human-auto-correlation} compares the AI judge with human experts. The AI judge rarely assigns \emph{Disagree} or \emph{Somewhat Disagree} for understanding and usability and slightly favors \name, giving it 28.6\% Agree rating in comparison to 14.3\% Agree ratings for \Baseline for understanding. However, for quality of choices, it rates both schemes poorly, with only 9.6\% \emph{Agree} or \emph{Somewhat Agree} for \name and 19\% for \Baseline.

\begin{figure}[h]
    \centering
    \begin{subfigure}{\linewidth}
        \centering
        \includegraphics[width=\linewidth]{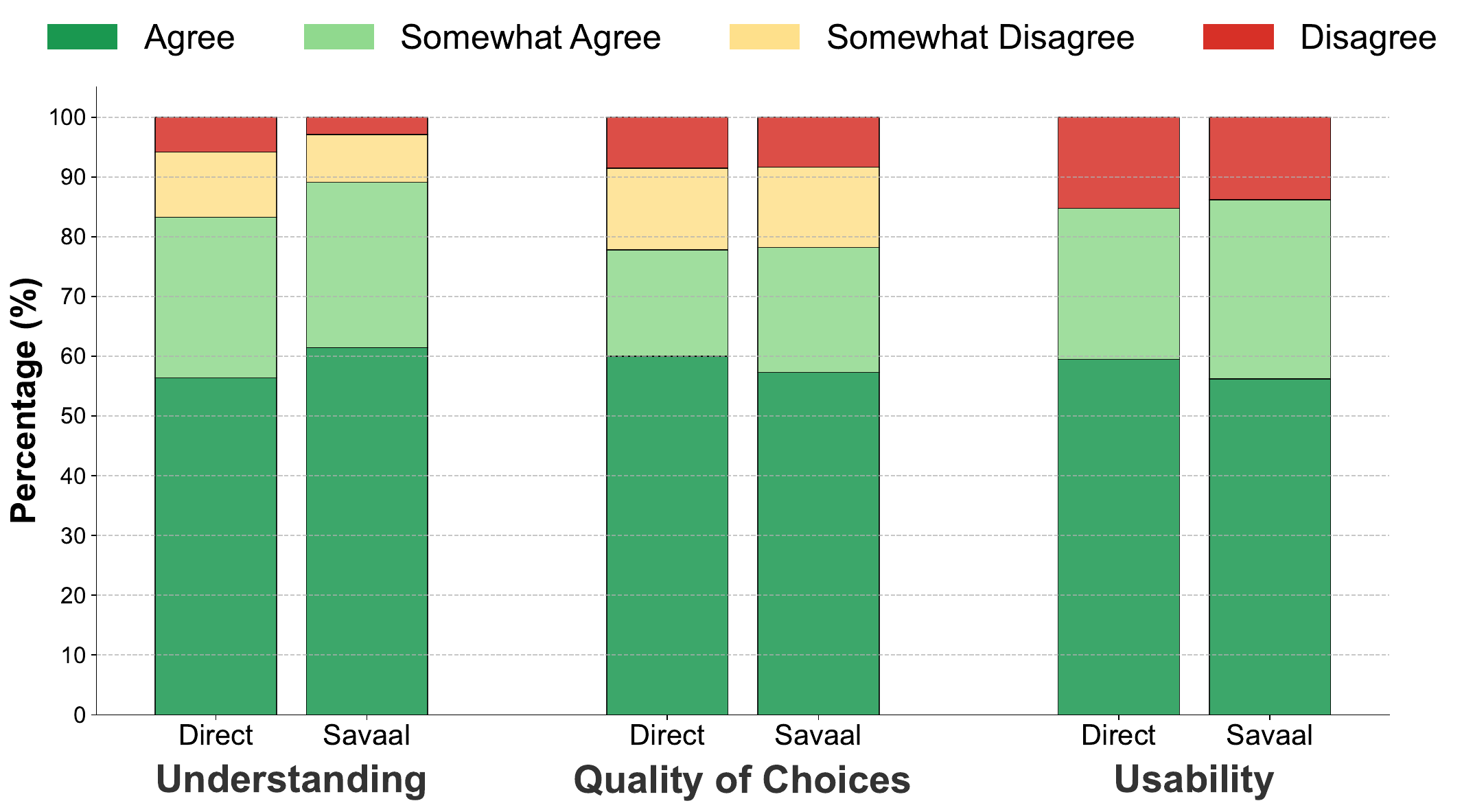}
        \caption{Breakdown of human expert scores on conference papers.}
        \label{fig:human_paper_breakdown}
    \end{subfigure}
    \hfill
    \begin{subfigure}{\linewidth}
        \centering
        \includegraphics[width=\linewidth]{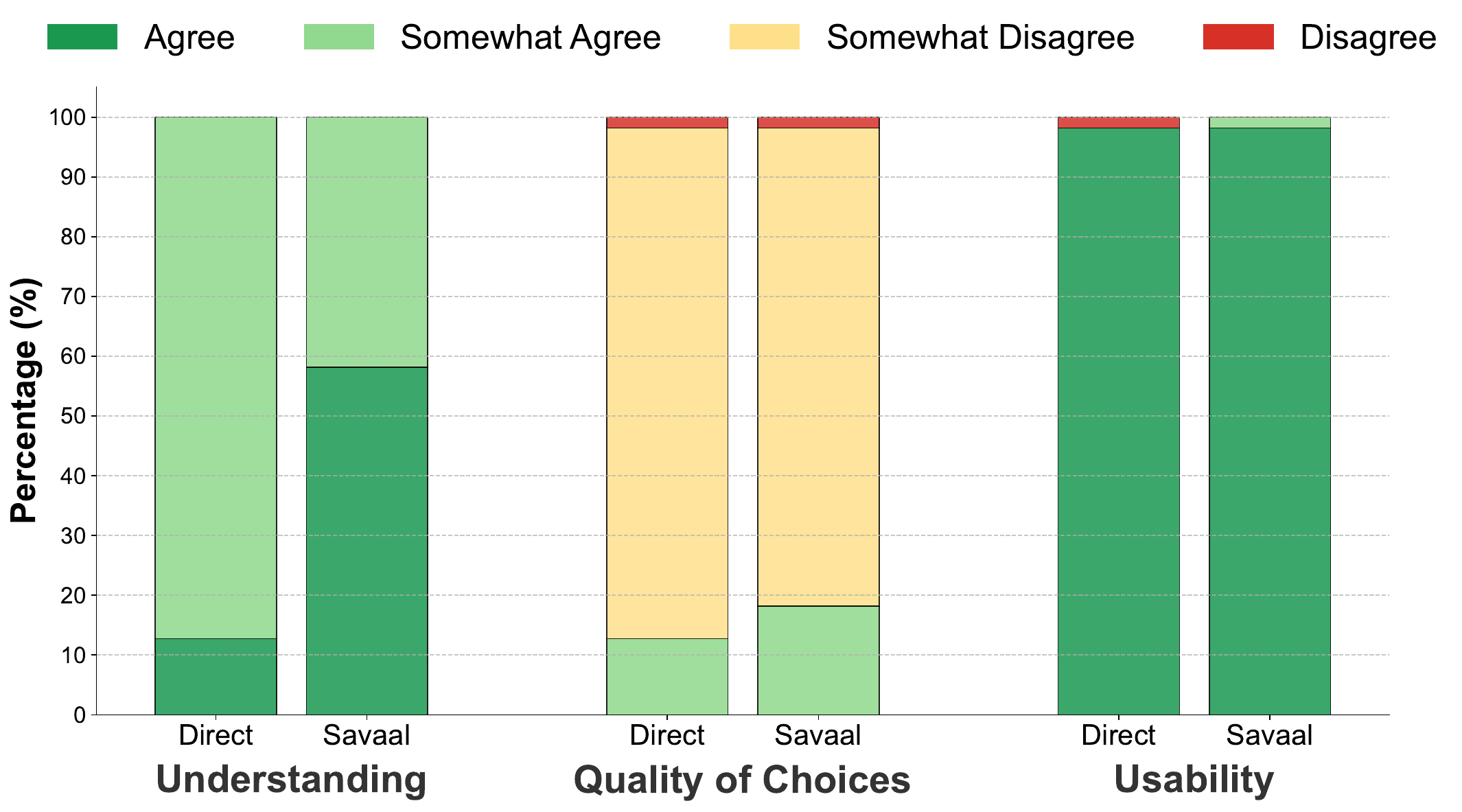}
        \caption{Breakdown of GPT-4o scores on conference papers.}
        \label{fig:AI_paper_breakdown}
    \end{subfigure}
    \caption{Score distribution for  \protect \numpaperquestions questions from conference papers.}
    \label{fig:paper_breakdown}
    \vspace{-10 pt}
\end{figure}

We observed similar trends in the \numpaperquestions questions from the conference-paper dataset (\autoref{fig:paper_breakdown}), where the AI judge again slightly preferred \name but remained misaligned with human expert evaluations. For completeness, we also present AI judge results on the Diverse \arxiv dataset in \autoref{subsec:ablation}.

%\paragraph{Limitations of the AI judge.} 
Our takeaway is that our \texttt{GPT-4o} AI judge was unaligned with human expert judgments (see \autoref{fig:auto-correlation-ai} vs. \autoref{fig:human-phd-breakdown}). Despite our extensive efforts in prompt engineering to maximize alignment---including using the prompt optimizer program in DSPy~\citep{khattab2024dspy}---AI-human correlation did not improve. Our experience calls into question the wisdom of using only AI judges in research studies.

\subsection{Cost Scalability}
\label{sec:scalability-case}

\begin{figure}[h]
    \centering
    \includegraphics[width=0.8\linewidth]{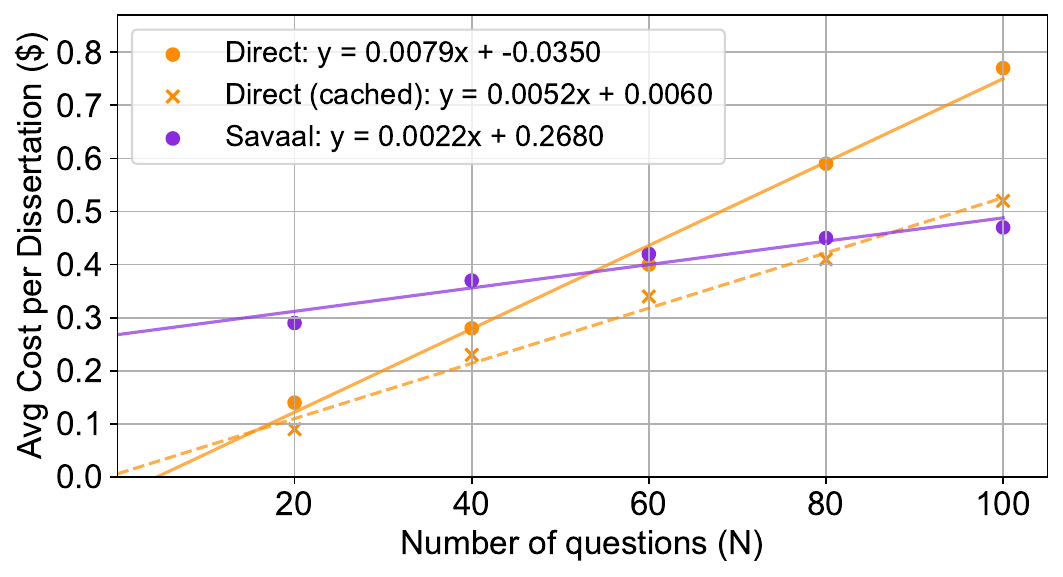}
    \caption{Average cost comparison of \BaselineMT and \name when generating questions from \numphd PhD dissertations. \name becomes less expensive as $N$ grows. We calculated costs by tracing prompt and completion tokens with OpenAI's February 2025 API pricing.}
    \label{fig:cost-scalability}
    \vspace{-20 pt}
\end{figure}

\autoref{fig:cost-scalability} compares the costs of \name and \BaselineMT on the dissertations. While \name incurs a higher one-time cost to generate the concepts, it becomes less expensive when generating more questions. At $N = 60$ questions, \name has the same cost as \BaselineMT; when $N$ grows to 100 questions, \BaselineMT is \directcostinflation more expensive. 

% Details are in \autoref{appendix:costs}.

\input{cost_detail}

%% file: cost_detail.tex
% \subsection{Discussion of Cost Scalability}
% \label{appendix:costs}

\name is also more cost-effective as the size of the document, $D$, grows. \BaselineMT costs $\approx \frac{N}{b} \cdot (A \cdot D + 100b \cdot B)$, where $A$ is cost per input token, $B$ is cost per output token, $N$ is the number of questions, $b$ is the batch size of \BaselineMT, and $100b$ is the approximate number of output tokens when generating $b$ questions. By contrast, \name costs $\approx f(D) + 100NB$ where $f(D)$ is the cost of main idea extraction, and $N$ is the number of questions. Thus, \name incurs a fixed cost that depends on the size of the document, but the marginal cost of generating additional questions is then independent of document size. By contrast, \Baseline incurs additional input token cost of $AD$ for each  batch of generated questions. 

In our experiments, for a PhD dissertation, $f(D) \approx 1.48A \cdot D$ on average.  Therefore, \name has lower cost when $\frac{N}{b} > 1.48$. For $N = 100$, \Baseline requires $b \approx 67$ to incur the same cost as \name, which is impractical with current LLMs. Both \texttt{GPT-4o} and \texttt{Meta-Llama-3.3-70B-Instruct} do not reliably generate more than $\approx$ 20 questions in a batch.

In \autoref{fig:cost-scalability}, we also notate \BaselineMT with caching. Prompt caching is a feature made available from various LLM providers. It works by matching a prompt prefix, like a long system prompt or other long context from previous multi-turn conversations, to reduce computation time and API costs. As of writing in February 2025, the OpenAI API charged 50\% less for cached prompt tokens, resulting in up-to 80\% latency improvements. The \BaselineMT method benefits from this caching scheme, as it repeatedly sends the entire document as a cache prefix to the API. As shown in \autoref{fig:cost-scalability}, \BaselineMT is more cost-effective than \name up until $N \approx 80$ with prompt caching, as opposed to $N \approx 60$ without prompt caching.

However, prompt caching has several limitations. First, many providers evict cache entries after a short period of time, around 5-10 minutes. Thus, all $N$ questions must be generated within a set time frame to benefit. Moreover, many open-source model providers do not include prompt caching as a feature (as of the time of writing). Therefore, while we present the benefits that prompt caching may provide \BaselineMT, we still demonstrate that \name is more cost effective at large scale.

%% file: 050related_work.tex
\section{Related Work}
\label{sec:related-work}

\textbf{Automated question-generation} has evolved from early Seq2Seq models \cite{du-etal-2017-learning, Neural_QG} to transformer-based approaches \cite{attention_is_all_you_need}. Models like BERT \cite{devlin-etal-2019-bert}, T5 \cite{t5}, BART \cite{lewis-etal-2020-bart}, and GPT-3 \cite{gpt_3} have significantly improved question generation \cite{chan-fan-2019-bert, li-etal-2021-addressing-semantic}. However, reliance on labeled datasets such as SQuAD \cite{rajpurkar-etal-2016-squad} and HotpotQA \cite{yang-etal-2018-hotpotqa} limits generalizability to other domains.

Researchers have explored LLMs for question generation~\cite{knowledge_base_prompting, reading_comprehension_language_llm, code_QG, mcq_computing, MIT_law}. However, these efforts have focused on generating questions from short, domain-specific context. Our work mitigates this limitation and generates high-quality questions from long documents.

Prior methods for \textbf{automated evaluation using LLMs} use metrics like ROUGE \cite{lin-2004-rouge} and BLEU \cite{papineni-etal-2002-bleu}, but these often misalign with humans~\cite{guo2024survey-neural-question-gen}. Some papers fine-tune small models for specific metrics \cite{zhu2023judgelm, wang2024pandalm}, but they face scalability issues, annotation reliance, or poor generalizability \cite{zhu2023judgelm}. Recent work uses  LLMs like GPT-4o as evaluators \cite{zheng2023judging, lin2023llmevalunifiedmultidimensionalautomatic}. While they achieve good human alignment, they focus on multi-turn conversations, a different context from ours.

For multiple-choice question generation, small models like BART and T5 assess relevance and usability \cite{moon-etal-2024-generative, raina2022multiplechoicequestiongenerationautomated} but require ground-truth data, limiting scalability. Others use LLM judges to rate relevance, coverage, and fluency on a 1-5 Likert scale \cite{microsoft_agriculture}. 
We adopt a similar approach with GPT-4o on a 1-4 scale. 
%for better human alignment. 
LLM judges can introduce positional \cite{zheng2024large, wang-etal-2024-large-language-models-fair}, egocentric \cite{koo-etal-2024-benchmarking}, and misinformation biases \cite{chen-etal-2024-humans, koo-etal-2024-benchmarking}.
%, which require mitigation.

\textbf{Retrieval-Augmented Generation (RAG)} enhances language model accuracy by retrieving relevant information to ground responses and reduce hallucinations~\cite{lewis2020retrieval, shuster2021retrieval, colbertv2, gottumukkala2022investigating}. Advances like dense passage retrieval~\cite{karpukhin2020dense} and late interaction models~\cite{colbert} improve efficiency. \name's pipeline uses recent advances in information retrieval models to fetch the most relevant context for question generation.

%% file: 060conclusions.tex
\section{Conclusion and Future Work}
\label{sec:concl}

\name uses LLMs and RAG in a concept-driven, three-stage framework to generate multiple-choice quizzes that assess deep understanding of large documents. Evaluations with \totalevaluators experts on \totaldocuments papers and dissertations show that, among those with a preference, \name outperforms a direct-prompting LLM baseline by 6.5$\times$ for dissertations and 1.5$\times$ for papers. Additionally, as document length increases, \name's advantages in question quality and cost efficiency become more pronounced.

We now discuss several avenues for future work.
While \name generates conceptual questions that test depth of understanding, few of them require mathematical analysis, logical reasoning, or creative thinking. \name produces quiz sessions, but we have not yet evaluated session quality. Currently, \name has not utilized human feedback to improve, which could be done using direct-preference optimization (DPO)~\cite{dpo}, Kahneman-Twersky Optimization (KTO) \cite{kto}, or reinforcement learning with human feedback (RLHF) \cite{rlhf}. To help learners, \name should adapt the difficulty of questions to the learner's answering accuracy and the time to answer questions.

Our attempts to align AI-generated evaluations with human expert judgments have been unsuccessful. Further research is necessary to improve AI judges in educational contexts.
Finally, validating \name's domain-independence requires testing across a broader spectrum of fields. 

%% file: limitations.tex
\section*{Limitations}

\NewPara{Number of human experts:} We presented results from \totalevaluators experts (authors). The number wasn't larger due to cost and time constraints. While we found that the quality of feedback is high and believe that this number is reasonable, it could be larger for greater statistical significance. Our hit rate on responses to the email invitations was \hitresponse, so there may have been some bias in who responded and completed the evaluation. We will continue to obtain more expert evaluations, but given our constraints, it is unlikely to be larger than a few hundred experts. 

\NewPara{Variety of domains:} \name is designed to be domain-independent, but as of now, we have evaluated it only in the areas of CS and Aero. However, our development has had no domain-specific engineering, training, or prompting. 

\NewPara{PDF document constraints:} This paper PDF documents parsed with GROBID, excluding figures from question generation. While our system supports web-based documents and follows hyperlinks, this paper evaluates only PDFs.

%\NewPara{Preference expression by human experts.} A more robust evaluation method could involve pairwise expert preferences, where evaluators compare two questions—one from each method—across multiple trials. This may provide a stronger ranking mechanism than absolute scoring.

%A better approach to ranking questions by quality might have been to have the expert choose a preference between two questions, one from each method, and run such an evaluation over $n$ pairs of questions. 

\NewPara{Session-level evaluation:} We evaluate individual questions but not full quiz sessions. Assessing entire quizzes is critical for measuring concept coverage and learning outcomes but is challenging due to {\em evaluator fatigue}.

\NewPara{Incorporating human feedback:} \name currently does not use any human feedback for fine-tuning or reinforcement learning. Doing so could enhance its quality and potentially improve other methods like \Baseline, altering the relative performance results reported.

%\jc{We don't do any human feedback tuning to fine-tune/RLHF/etc. the model, which we should do for future work}

\NewPara{Question types:} This paper focuses on single-answer multiple-choice questions, though real-world tests use diverse formats, including multiple-correct-choice, true/false, fill-in-the-blank, and open-ended questions. Currently, \name generates high-quality conceptual questions (as shown by our results), but does not yet produce ones requiring logical or mathematical reasoning.

% \pk{@Hari, should we add option refinement here as well? }

%\kn{From ACL website: Authors are required to discuss the limitations of their work in a dedicated section titled “Limitations”. This section should be included at the end of the paper, before the references, and \textbf{it will not count toward the page limit.} This includes both, long and short papers. Papers without a limitations section will be \textbf{desk rejected.}}

\section*{Ethical Considerations}

Using LLMs to generate questions raises important ethical concerns regarding their responsible use in the training and education of people~\cite{MIT_law}. LLMs suffer from bias caused by their training data~\cite{bender2021dangers}, which can affect the quality and neutrality of the generated questions.

We conform to the ACL Code of Ethics~\cite{ACLCodeOfEthics}. Prior to our evaluation study, we obtained an IRB exemption. We have protected the privacy and anonymity of the evaluators by sharing only aggregate, anonymized statistics. The responses from our evaluators carry no risk of harm. Before participating, all evaluators reviewed a consent form and provided feedback through a secure platform (see~\autoref{subsec:appendix_human_eval_conduct} for details). We use the term ``expert'' to refer to an author of the evaluated documents, but this label does not imply any specific responsibilities or expectations on the evaluator. All evaluators took part voluntarily, without compensation.

We envision \name to help learners and educators by generating questions. It is not intended to replace human teachers. LLMs are prone to errors and hallucinations and may learn biased information from training data~\cite{MIT_law}. Therefore, an expert or educator needs to ensure that the questions and answers generated by \name are accurate and relevant to the material.  

Generating questions from research papers introduces potential concerns regarding intellectual property, copyright, and attribution. \name does not copy text directly from documents but synthesizes questions based on inferred key concepts. Users should acknowledge original sources when using \name, particularly in educational, research, and commercial settings.

%% file: 070appendix.tex
\clearpage
\appendix

\section{Appendix}

\input{human_results_detail}
\input{012ablation}

\subsection{Details of Conducting the Expert Study}
\label{subsec:appendix_human_eval_conduct}

\begin{figure*}[h]
\centering
\includegraphics[width=0.85\linewidth]{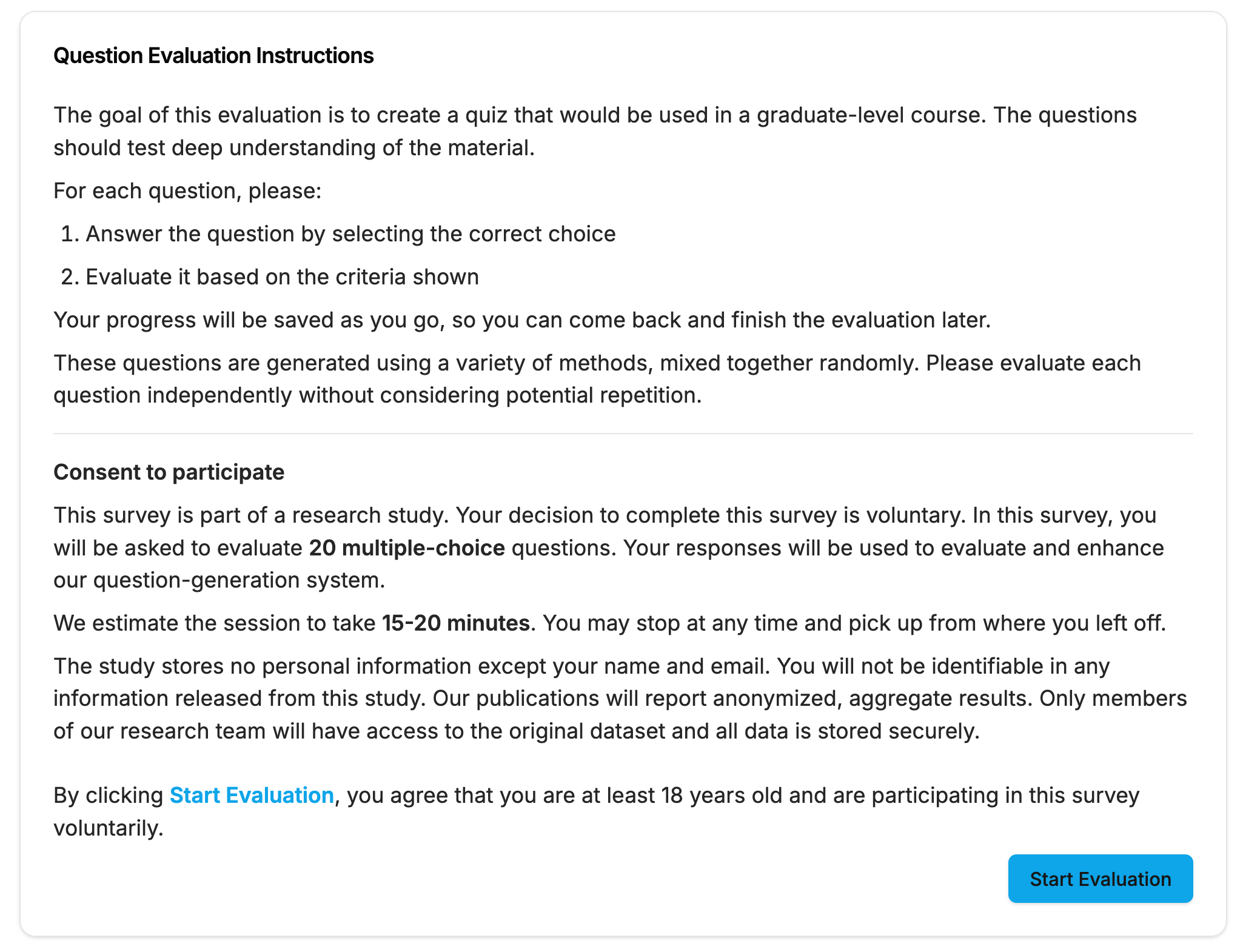}
    \caption{Consent form for the human evaluation}
    \label{fig:savaal-consent}
\end{figure*}

\begin{figure*}[h]
\centering
\includegraphics[width=0.8\linewidth]{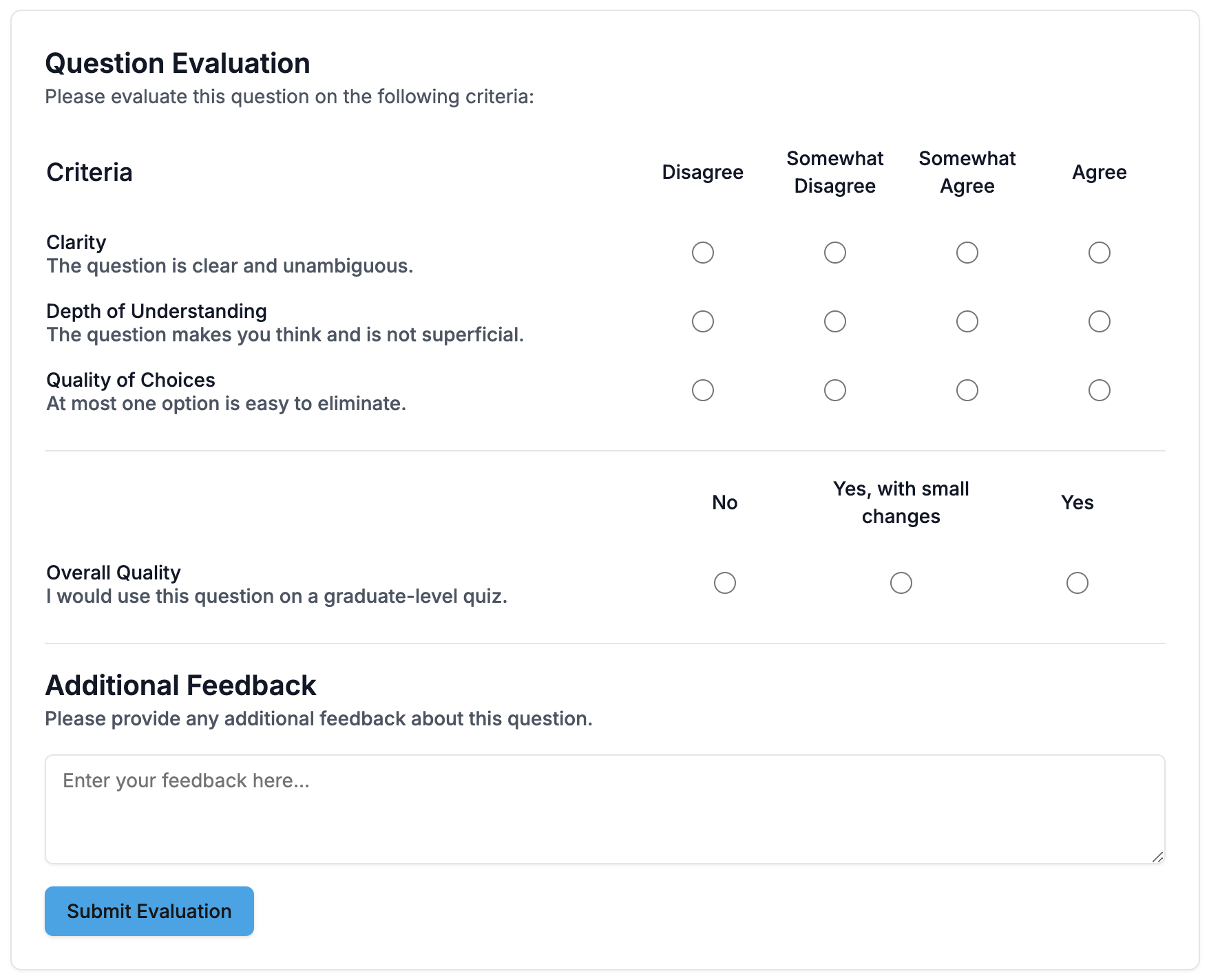}
    \caption{Form for the expert evaluations.}
    \label{fig:savaal-form}
\end{figure*}

To conduct the human evaluation, participants were first required to review and sign a consent form that outlined the study's purpose, data privacy, and the voluntary nature of their participation (\autoref{fig:savaal-consent}). After signing the consent form, participants completed a blind evaluation form consisting of 20 randomly selected questions from \name and \Baseline. They assessed each question based on clarity, depth of understanding, quality of choices, and overall usability (\autoref{fig:savaal-form}). All responses were anonymized, and participants had the option to withdraw from the study at any time.

\subsection{Prompts}
\label{subsec:appendix_prompts}
\input{080prompts}
\input{100refiner}

\subsection{Examples}
\label{subsec:appendix_examples}
\input{090examples}

%% file: human_results_detail.tex
\subsection{Observations from Expert Evaluations}
\label{subsec:human_eval_observations}
We discuss some additional findings from our expert evaluations. 

\subsubsection{Bias When Responding Incorrectly}
\label{subsubsec:human_eval_observations_bias}

\begin{figure}[h]
    \centering
    \begin{subfigure}{\columnwidth}
        \centering
        \includegraphics[width=\linewidth]{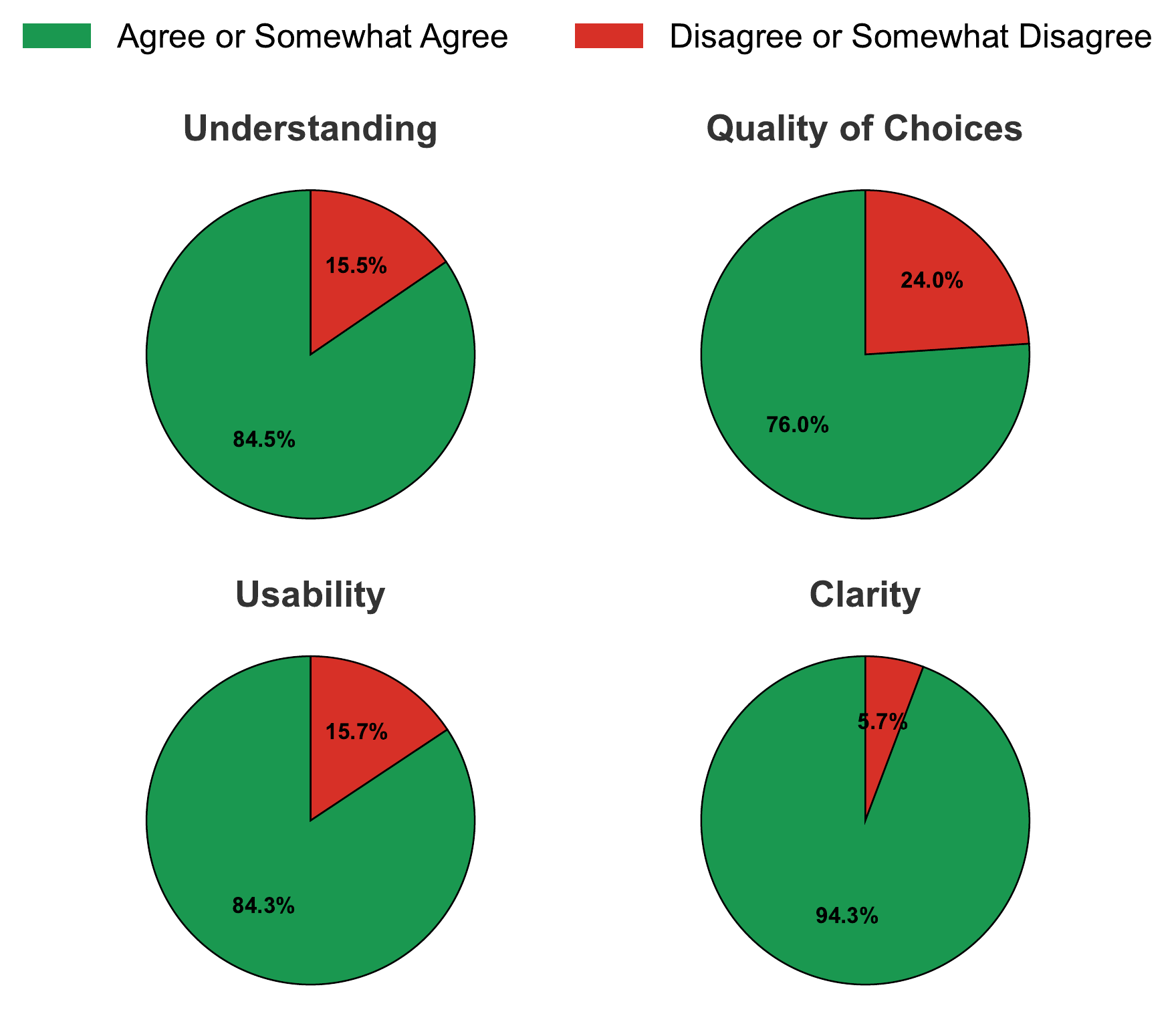}
        \caption{Ratings for \textbf{correct} responses (1411 questions).}
        \label{fig:correct-circle}
    \end{subfigure}
    \hfill
    \begin{subfigure}{\columnwidth}
        \centering
        \includegraphics[width=\linewidth]{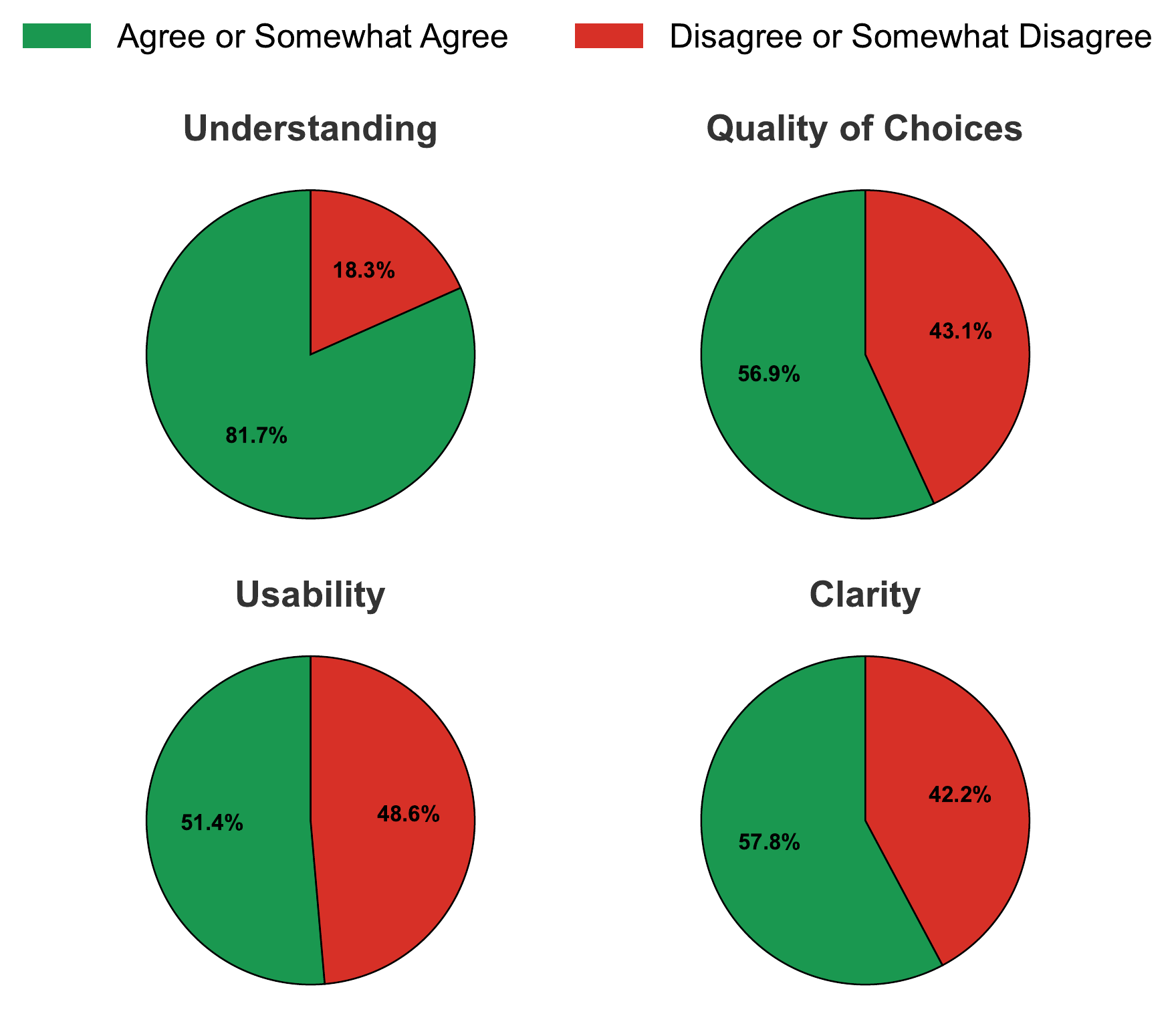}
        \caption{Ratings for \textbf{incorrect} responses (109 questions).}
        \label{fig:incorrect-circle}
    \end{subfigure}
    \caption{Comparison of expert ratings on different metrics for correct and incorrectly answered questions.}
    \label{fig:savaal-circles}
\end{figure}

Prior to rating a question, evaluators select a response and see the ``correct'' answer (more accurately, the choice that the question generation system thinks is correct).  Experts rate questions that they answer ``correctly'' differently from those that they answer incorrectly. \autoref{fig:correct-circle} shows the distribution of responses across 1411 correctly answered questions (695 \name and 716 \Baseline), while \autoref{fig:incorrect-circle} shows the same for 109 questions answered incorrectly (65 \name and 44 \Baseline). When experts select the wrong answer, they penalize the quality of choices, usability, and clarity. However, their rating for depth of understanding is relatively unaffected. 

% \begin{figure*}[h]
%     \centering
%     \begin{subfigure}{0.44\linewidth}
%         \centering
%         \includegraphics[width=\linewidth]{FIG/all_correct_circle_metrics.pdf}
%         \caption{Ratings for \textbf{correct} responses (1411 questions).}
%         \label{fig:correct-circle}
%     \end{subfigure}
%     \hfill
%     \begin{subfigure}{0.44\linewidth}
%         \centering
%         \includegraphics[width=\linewidth]{FIG/all_incorrect_circle_metrics.pdf}
%         \caption{Ratings for \textbf{incorrect} responses (109 questions).}
%         \label{fig:incorrect-circle}
%     \end{subfigure}
%     \caption{Comparison of expert ratings on different metrics for correct and incorrectly answered questions.}
%     \label{fig:savaal-circles}
% \end{figure*}

\subsubsection{Inter-Human Correlation}

\begin{figure*}[h]
    \centering
    \begin{subfigure}[t]{0.27\linewidth}
        \centering
        \includegraphics[width=\linewidth]{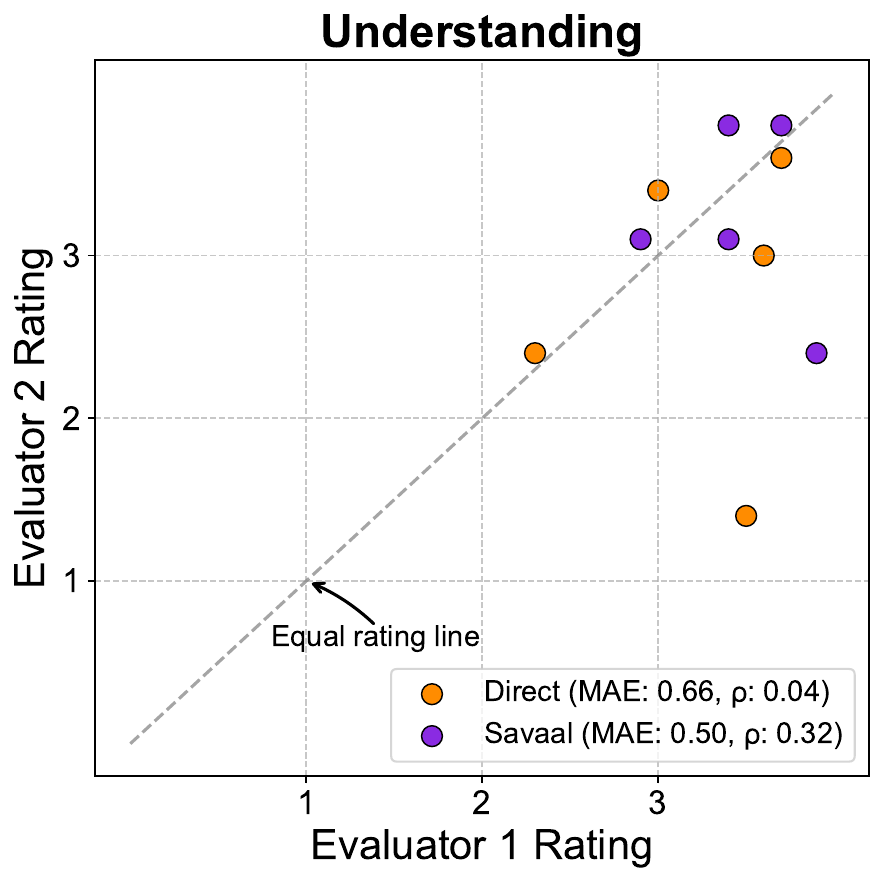}
        \caption{Human correlation on depth of understanding. Both \name and \Baseline exhibit weak correlation ($\rho = 0.32$ and $\rho = 0.04$ respectively).}
        \label{fig:correlation-scatter-understanding}
    \end{subfigure}
    \hfill
    \begin{subfigure}[t]{0.27\linewidth}
        \centering
        \includegraphics[width=\linewidth]{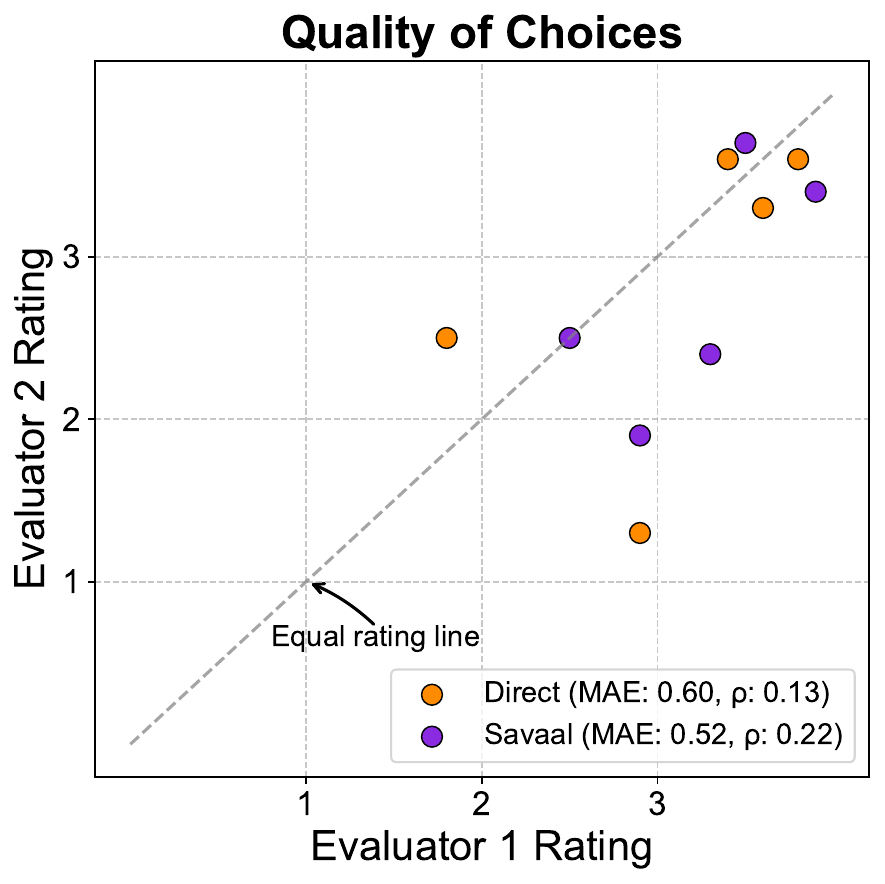}
        \caption{Human correlation on quality of choices. Both \name and \Baseline exhibit weak correlation ($\rho = 0.22$ and $\rho = 0.13$ respectively).}
        \label{fig:correlation-scatter-choices}
    \end{subfigure}
    \hfill
    \begin{subfigure}[t]{0.27\linewidth}
        \centering
        \includegraphics[width=\linewidth]{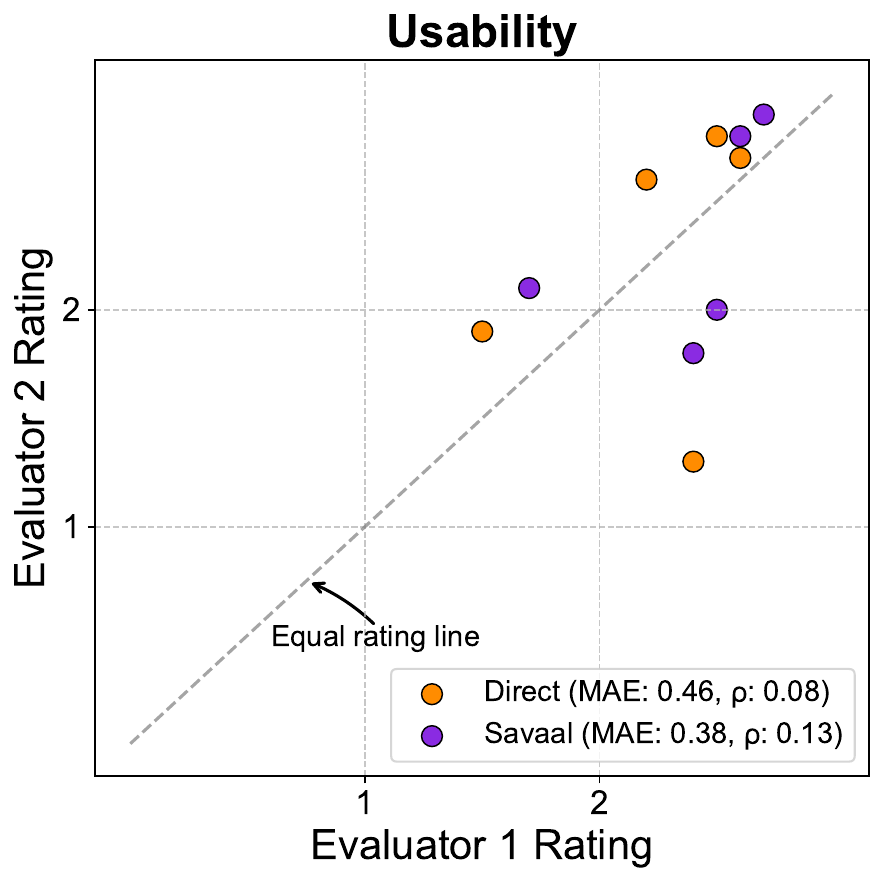}
        \caption{Human correlation on usability. Both \name and \Baseline exhibit weak correlation ($\rho = 0.13$ and $\rho = 0.08$ respectively).}
        \label{fig:correlation-scatter-usability}
    \end{subfigure}
    \caption{Correlation between human evaluators on the same document across metrics. Each point is the score of Evaluator 1 vs. Evaluator 2 on a particular document. $y = x$ is where human evaluators perfectly align with each other. We also compute the Mean Average Error (MAE), as well as the average Spearman correlation coefficient $\rho$.}
    \label{fig:inter-human-correlation}
\end{figure*}

On the conference paper dataset, there were 5 papers with two evaluators each. We examine the correlation of their scores in \autoref{fig:inter-human-correlation}. Each point represents Evaluator 1's average score compared to Evaluator 2's average score across each metric. We plot against the perfect-agreement $y = x$ line. To quantify their differences, we also compute the Mean Absolute Error (MAE) for each method across all pairs of evaluators and the average Spearman coefficient, which is measured on the pairwise ordinal observations on each question per document, averaged across the methods.

\begin{figure*}[h]
    \centering
    \begin{subfigure}[t]{0.27\linewidth}
        \centering
        \includegraphics[width=\linewidth]{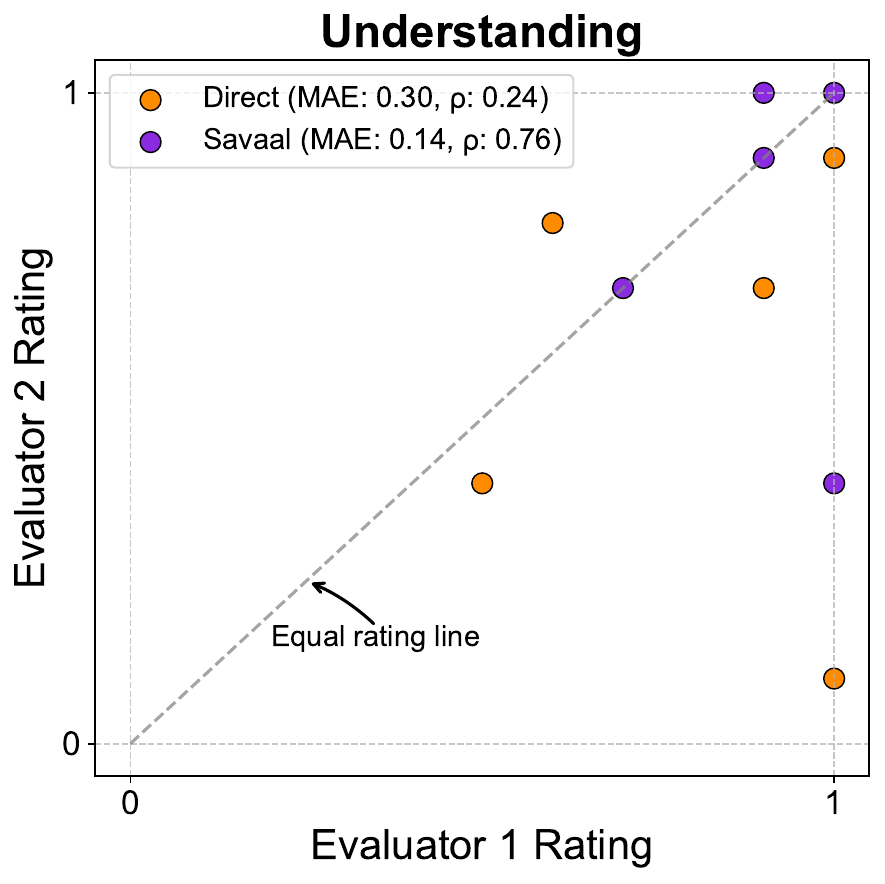}
        \caption{Human correlation on binarized depth of understanding. \name shows strong correlation ($\rho = 0.76$) while the \Baseline shows weak correlation ($\rho = 0.24$).}
        \label{fig:bin-correlation-scatter-understanding}
    \end{subfigure}
    \hfill
    \begin{subfigure}[t]{0.27\linewidth}
        \centering
        \includegraphics[width=\linewidth]{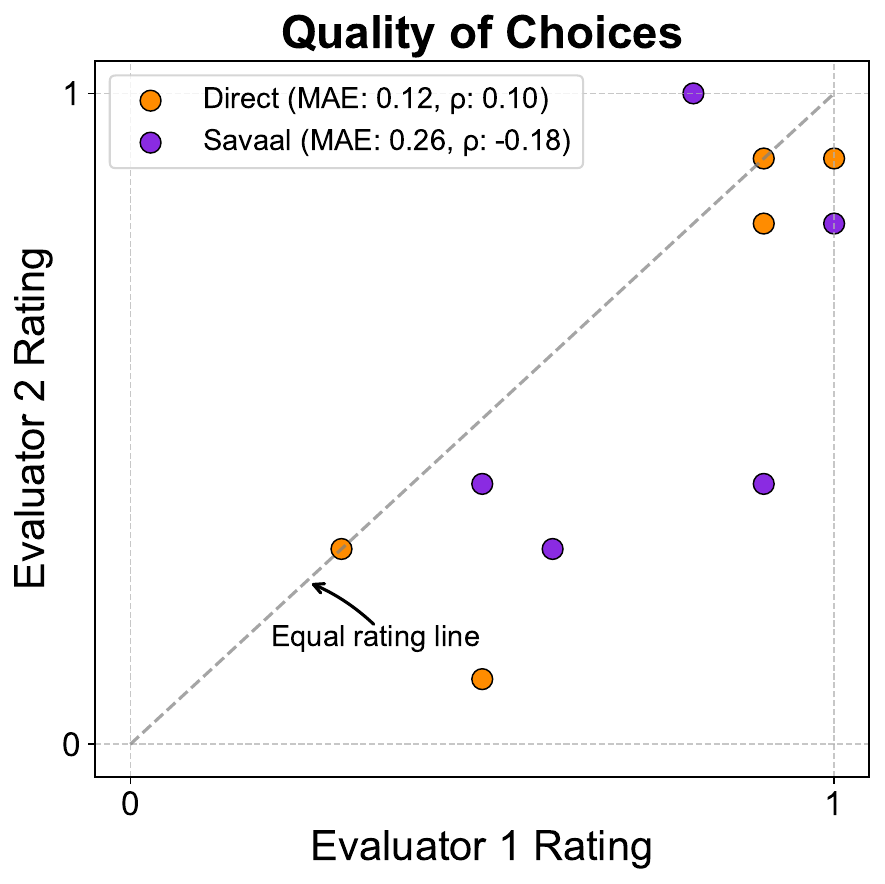}
        \caption{Human correlation on binarized quality of choices. \Baseline showed weak correlation ($\rho = 0.10$) while \name showed negative weak correlation ($\rho = -0.18$).}
        \label{fig:bin-correlation-scatter-choices}
    \end{subfigure}
    \hfill
    \begin{subfigure}[t]{0.27\linewidth}
        \centering
        \includegraphics[width=\linewidth]{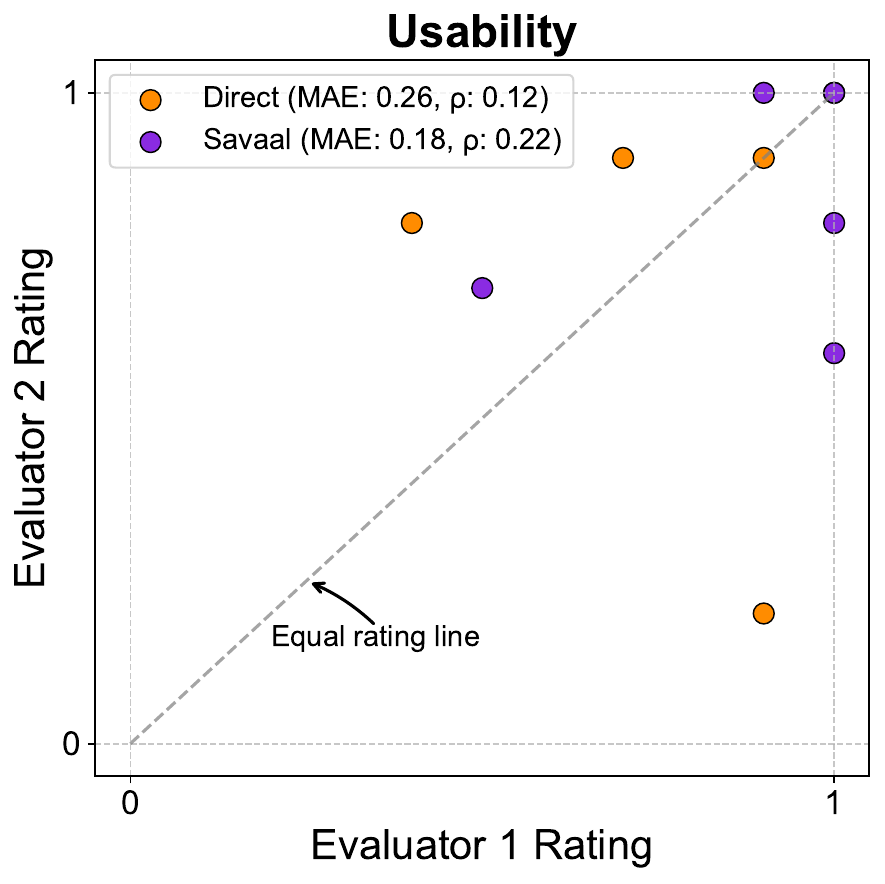}
        \caption{Human correlation on binarized usability. Both \name and \Baseline exhibit weak correlation ($\rho = 0.22$ and $\rho = 0.12$ respectively).}
        \label{fig:bin-correlation-scatter-usability}
    \end{subfigure}

    \caption{Correlation between human evaluators on the same document across metrics. Each point is the score of Evaluator 1 vs. Evaluator 2 on a particular document. $y = x$ is where human evaluators perfectly align with each other. We also compute the Mean Average Error (MAE), as well as the average Spearman correlation coefficient $\rho$.}
    \label{fig:binary-inter-human-correlation}
\end{figure*}

We find that the evaluators had poor correlation between themselves when visualizing their aggregate scores for each method (\autoref{fig:inter-human-correlation}). Binarizing their scores, however, increased their correlations, particularly for depth of understanding ($\rho = 0.76$) (\autoref{fig:binary-inter-human-correlation}). We expect that with more samples of evaluations drawn from the same set of questions, we can find stronger correlation trends.

%% file: 012ablation.tex
.
\clearpage

\subsection{Additional Methods and Quality Criteria}
\label{subsec:ablation}

We extend the evaluation to compare \name against other methods and metrics using the AI judge on the arXiv dataset. For these experiments, we generate 20 questions per method for each paper. 

\autoref{tab:benchmark-stats} provides further information  about the arXiv dataset. It consists of 48 scientific papers across five topic categories: Computer Science, Physics, Mathematics, Economics, and Quantitative Biology. These papers are divided into two sub-categories: \emph{old} and \emph{new}.

\begin{itemize}[topsep=0pt, itemsep=0pt, leftmargin=*]
    \item \emph{new} Papers: papers published on arXiv after October 2023, which is after the knowledge cutoff date for the LLMs used in this paper. We randomly selected five papers per category from arXiv.
    \item \emph{old} Papers: papers published on arXiv prior to October 2023. We randomly selected five papers per category from the LooGLE dataset~\cite{loogle}, except for Quantitative Biology, where only three papers were available on LooGLE.
\end{itemize}

We split the dataset into ``old'' and ``new'' papers to evaluate whether the performance is different on documents that were not included in the LLM's training data. We did not observe any significant differences for old and new papers, with any of the question generation methods. Thus, we aggregate results for old and new papers for the analysis below.

\begin{table*}[t]
\centering
\renewcommand{\arraystretch}{1.2} % Increases row height for better readability
\setlength{\tabcolsep}{8pt} % Adjusts column spacing
\resizebox{\textwidth}{!}{%
\begin{tabular}{|c|cc|cc|cc|cc|cc|}
\hline
\multirow{2}{*}{\textbf{Category}} & \multicolumn{2}{c|}{\textbf{Computer Science}} & \multicolumn{2}{c|}{\textbf{Physics}} & \multicolumn{2}{c|}{\textbf{Mathematics}} & \multicolumn{2}{c|}{\textbf{Economics}} & \multicolumn{2}{c|}{\textbf{Quantitative Biology}} \\ 
\cline{2-11} 
 & \multicolumn{1}{c|}{\textbf{Old}} & \textbf{New} & \multicolumn{1}{c|}{\textbf{Old}} & \textbf{New} & \multicolumn{1}{c|}{\textbf{Old}} & \textbf{New} & \multicolumn{1}{c|}{\textbf{Old}} & \textbf{New} & \multicolumn{1}{c|}{\textbf{Old}} & \textbf{New} \\ 
\hline
\textbf{No. Papers} & \multicolumn{1}{c|}{5} & 5 & \multicolumn{1}{c|}{5} & 5 & \multicolumn{1}{c|}{5} & 5 & \multicolumn{1}{c|}{5} & 5 & \multicolumn{1}{c|}{3} & 5 \\ 
\hline
\multirow{2}{*}{\textbf{Avg. Words}} & \multicolumn{1}{c|}{12,498} & 7,307 & \multicolumn{1}{c|}{14,298} & 21,088 & \multicolumn{1}{c|}{12,049} & 16,596 & \multicolumn{1}{c|}{14,010} & 16,112 & \multicolumn{1}{c|}{19,390} & 6,613 \\ 
\cline{2-11} 
 & \multicolumn{2}{c|}{9,903} & \multicolumn{2}{c|}{17,693} & \multicolumn{2}{c|}{14,323} & \multicolumn{2}{c|}{15,061} & \multicolumn{2}{c|}{11,404} \\ 
\hline
\end{tabular}%
}
\caption{Statistics for the number of words for the random papers selected for Diverse \arxiv dataset.}
\label{tab:benchmark-stats}
\end{table*}

\paragraph{Additional Comparison Methods:}
\label{subsec:ablation-methods}
In addition to \Baseline (\autoref{sec:evaluation-baselines}), we consider two other strategies:
\begin{itemize}[topsep=1pt, itemsep=0pt, leftmargin=*]

    \item \textbf{Summary}: Uses the summary of the document as the context for question generation (\autoref{sec:insights-summary}). The summary is generated using a map-reduce approach. The prompt used to generate questions from the summary is identical to the \Baseline prompt (\autoref{fig:baseline_question_generation_prompt}).

    \item \textbf{Single-Prompt \TheSystem}: Concatenate all of the prompts used in the stages of \name's pipeline (\autoref{sec:pipeline}) into a single prompt, using the entire document as context. We described each step of \name's pipeline (see \autoref{fig:savaal-workflow}) in detail, and asked the LLM to ``think step by step'' and follow the steps (prompt not shown due to its long length).
    
\end{itemize}

\paragraph{Additional Metrics:}
\label{subsec:ablation-metrics}
In addition to Understanding, Quality of Choices, and Usability, we consider additional criteria for the AI judge to evaluate the questions. These metrics include difficulty, cognitive level, and engagement. The prompts used for these criteria are shown in \autoref{subsubsec:eval-prompts}.

\begin{figure*}[t]
    \centering
    \begin{subfigure}{\linewidth}
        \centering
        \includegraphics[width=\linewidth]{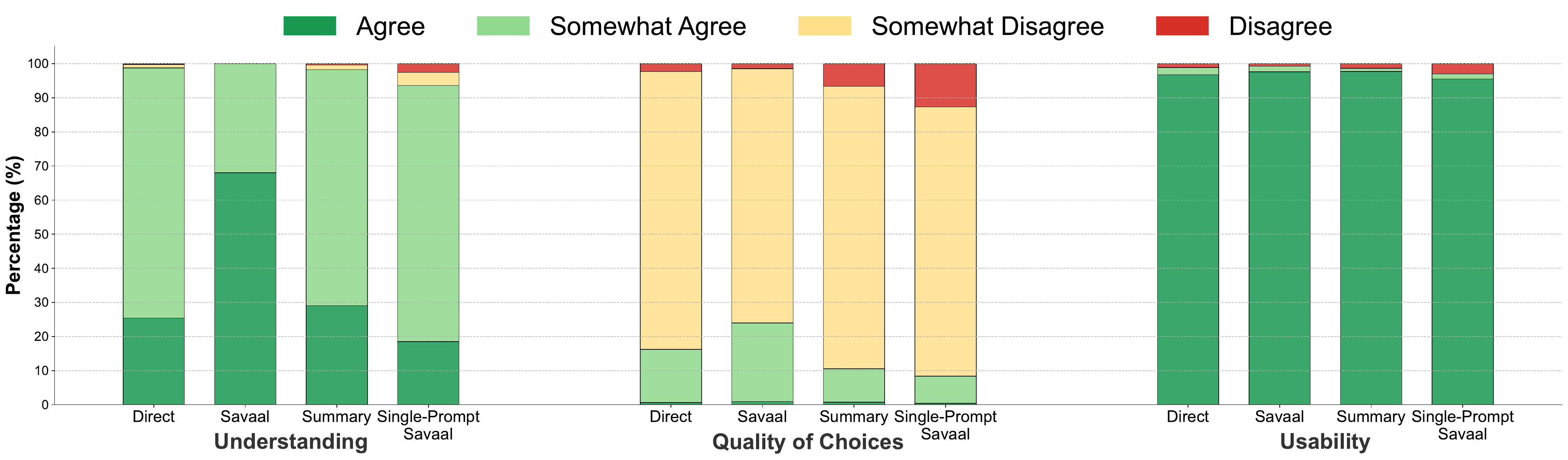}
        % \caption{Evaluation of GPT-4o on primary metrics.}
        \label{fig:auto_gpt_primary}
    \end{subfigure}
    \begin{subfigure}{\linewidth}
        \centering
        \includegraphics[width=\linewidth]{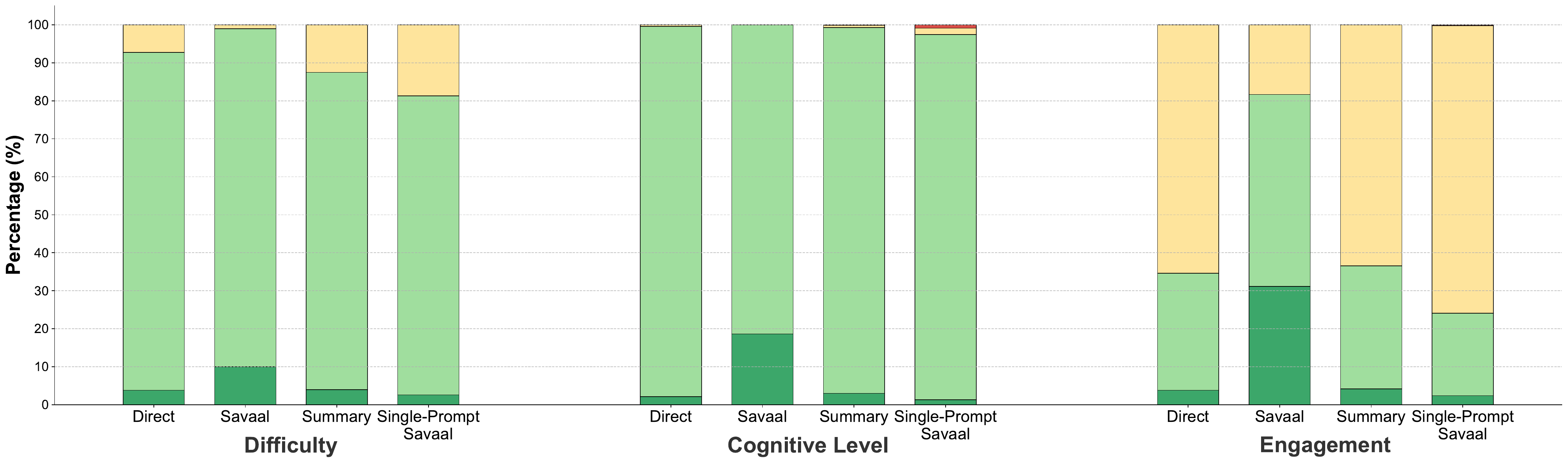}
        \label{fig:auto_gpt_additional}
    \end{subfigure}
    \caption{Results of AI evaluation on the quizzes generated with \texttt{GPT-4o} on the \arxiv dataset, evaluated by the AI Judge (\texttt{GPT-4o}).}
    \label{fig:auto_gpt_metrics}
\end{figure*}

\paragraph{Results:} \autoref{fig:auto_gpt_metrics} summarizes the AI judge scores on all metrics (Understanding, Quality of Choices, Usability, Difficulty, Cognitive Level, Engagement) across all methods (\autoref{subsec:ablation-methods}). The judge rates most of the questions with any method as usable, with the highest amount of usability for \name's questions. It also does not rate any method highly in terms of quality of choices, but gives \name the highest percentage of \emph{Agree}s and the lowest percentage of \emph{Disagree}s among all the methods. On the other criteria, Savaal performs better according to the AI judge.

\begin{figure}[t]
        \centering
            \includegraphics[width=\linewidth]{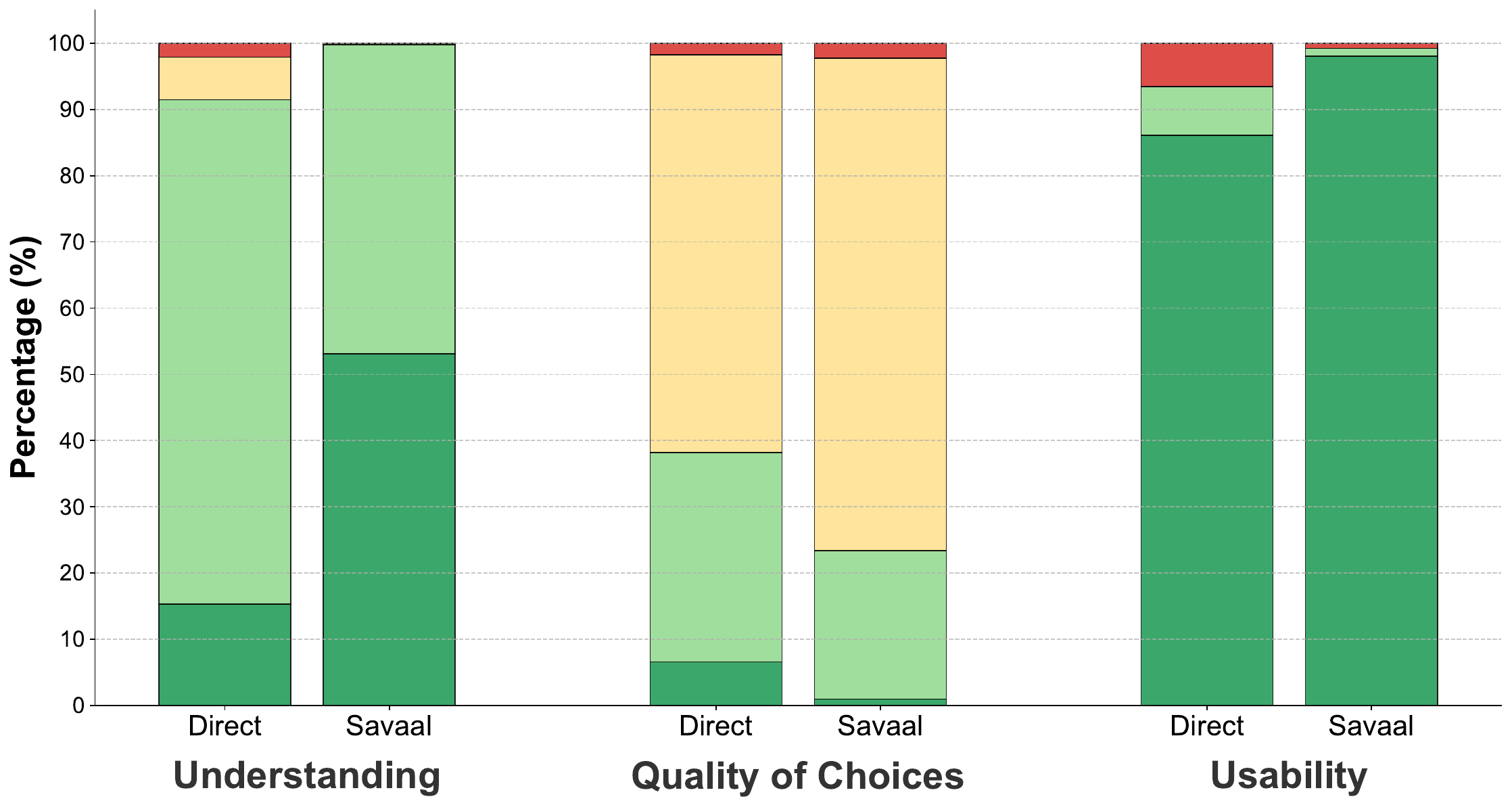}
        \caption{Results of AI evaluation on the quizzes generated with Llama-3.3-70B on the \arxiv dataset evaluated by the AI Judge (\texttt{GPT-4o}).}
        \label{fig:auto_llama_metrics}
    \end{figure}

\subsection{Evaluating \name with other LLMs}
\label{subsec:ablation-model}
To understand the sensitivity of our results to the underlying LLM, we replace \texttt{GPT-4o} with \texttt{Meta-Llama-3.3-70B-Instruct} and generate questions using the different methods. We used model \texttt{Meta-Llama-3.3-70B-Instruct} hosted at Together.ai~\citep{together_ai_api} for these experiments. We use \texttt{GPT-4o} as the AI judge for these experiments.

\autoref{fig:auto_llama_metrics} shows the scores that the AI judge gives to the questions generated using the Llama-3.3-70B-Instruct model with \Baseline and \name. The trends are similar for Llama-generated and GPT-4o questions: the AI judge rates \name higher in terms of depth of understanding and usability. It rates both \Baseline and \name poorly on choice quality overall, but prefers \Baseline for Llama-generated questions.

%% file: 080prompts.tex
\subsubsection{Question Generation Prompts}
\label{subsubsec:question-generation-prompts}

\autoref{fig:baseline_question_generation_prompt} presents the \Baseline question generation prompt. \BaselineMT builds upon this by generating additional unique questions, as shown in \autoref{fig:baseline_large_question_generation_prompt}. Similarly, \autoref{fig:question_generation} introduces the \name question generation prompt, used in step \circled{5}{customorange} of \autoref{fig:savaal-workflow}, which closely resembles the \Baseline prompt. Beyond question generation, \autoref{fig:map_prompt} depicts the map prompt from step \circled{1}{customblue}, while \autoref{fig:combine_prompt} and \autoref{fig:reduce_prompt} (step \circled{2}{customred}) extend this by consolidating multiple concept maps into a comprehensive summary. Finally, \autoref{fig:main_idea_ranking_prompt} illustrates the ranking prompt used in step \circled{3}{custompurple} of \autoref{fig:savaal-workflow}.

%%%%%%%%%%%%%%%%%%%%%% Question Generation Prompt %%%%%%%%%%%%%%%%%%%%%%

\begin{figure}[h]
\centering
\begin{tcolorbox}[width=1\columnwidth, fontupper=\footnotesize,title=\Baseline Question Generation Prompt]
\RaggedRight
\obeylines
\textbf{Instructions}: \\
{
Based on the following context, create $\{\text{num\_questions}\}$ multiple-choice questions that require deep understanding, critical thinking, and detailed analysis.  
The questions should go beyond mere factual recall, involving higher-order thinking skills like analysis, synthesis, and evaluation.

\textbf{Provide four answer choices for each question:}
\begin{itemize}[label=-,nosep]
    \item The choices should start with \textbf{A.}, \textbf{B.}, \textbf{C.}, and \textbf{D.}
    \item \textbf{One correct answer.}
    \item \textbf{Three plausible distractors} that are:
    \begin{itemize}[label=$\bullet$,nosep]
        \item Contextually appropriate.
        \item Relevant to the content.
        \item Reflect common misunderstandings or errors without introducing contradictory or irrelevant information.
    \end{itemize}
\end{itemize}
\vspace{3mm}
\textbf{Note:} The questions should focus on one concept and not be overly long.  
\textbf{DO NOT} ask multiple questions in one.
}
\vspace{3mm}
\vspace{3mm}
\textbf{Context}: \\
$\{\text{context}\}$ \\

\end{tcolorbox}
\caption{\Baseline Question Generation Prompt.}
\label{fig:baseline_question_generation_prompt}
\end{figure}

%%%%%%%%%%%%%%%%%%%%%% Question Generation %%%%%%%%%%%%%%%%%%%%%%

 \begin{figure}[h]
\centering
\begin{tcolorbox}[width=1\columnwidth, fontupper=\footnotesize,title= Savaal Question Generation Prompt]
\RaggedRight
\obeylines
\textbf{Instructions}: \\
{Based on the following main idea and its relevant passages, create $\{\text{num\_questions}\}$ \\ multiple-choice questions that require deep understanding, critical thinking, and detailed analysis. The questions should go beyond mere factual recall, involving higher-order thinking skills like analysis, synthesis, and evaluation. \\
Do not use the phrases "main idea" or "passages" in the question statement. Instead, directly address the content or concepts described. \\
Provide four answer choices for each question: \\
\begin{itemize}[label=-,nosep]
    \item The choices should start with A., B., C., and D.
    \item One correct answer.
    \item \textbf{Three plausible distractors} that are contextually appropriate, relevant to the content, and reflect common misunderstandings or errors without introducing contradictory or irrelevant information.
\end{itemize}

\vspace{3mm}
\textbf{Note}: The questions should be focused on one concept and not very long, DO NOT ask multiple questions in one.} \\
\vspace{3mm}
\vspace{3mm}
\textbf{Main Idea}: \\
$\{\text{main\_idea}\}$ \\
\vspace{3mm}
\textbf{Passages}: \\
$\{\text{passages}\}$ \\

\end{tcolorbox}
\centering
\caption{The question generation prompt in \autoref{fig:savaal-workflow}.}
\label{fig:question_generation}
\end{figure}

% \vspace{-20 pt}
%%%%%%%%%%%%%%%%%%%%%% MAP %%%%%%%%%%%%%%%%%%%%%%

\begin{figure}[h]
\centering
\begin{tcolorbox}[width=1\columnwidth, fontupper=\footnotesize,title=Map Prompt]
\RaggedRight
\obeylines
\textbf{Instructions}: \\
{You are an expert educator specializing in creating detailed concept maps from academic texts. Given the following excerpt from a longer document, extract the main ideas, detailed concepts, and supporting details that are critical to understanding the material. \vspace{3mm}
Focus on identifying:
\begin{itemize}[label=-,nosep]
    \item Key concepts or terms introduced in the text.
    \item Definitions or explanations of these concepts.
    \item Relationships between concepts.
    \item Any examples or applications mentioned.
\end{itemize}
\vspace{3mm}
\vspace{3mm}
Use clear, bullet-point summaries, organized by topic. Here is the excerpt:}
\vspace{3mm}
\vspace{3mm}
\textbf{Context}: \\
$\{\text{context}\}$ \\

Respond with a structured list of detailed main ideas and concepts.
\end{tcolorbox}
\caption{\centering The map prompt in \autoref{fig:savaal-workflow}.}
\label{fig:map_prompt}
\end{figure}

%%%%%%%%%%%%%%%%%%%%%% Combine %%%%%%%%%%%%%%%%%%%%%%
\begin{figure}[h]
\centering
\begin{tcolorbox}[width=1\columnwidth, fontupper=\footnotesize,title=Combine Prompt]
\RaggedRight
\obeylines
\textbf{Instructions}: \\
{You are combining multiple concept maps into a single, comprehensive summary while retaining all key ideas and details. Below are several lists of main ideas and concepts extracted from a larger document.%\\ \\
\vspace{3mm}
Your task is to:
\begin{enumerate}[nosep]
    \item Merge these lists into a single structured list, removing redundancies while keeping all unique and detailed information.
    \item Ensure all main ideas, relationships, and examples are preserved and clearly organized.
\end{enumerate}
\vspace{3mm}
\vspace{3mm}
Here are the concept maps to combine:} \\
\vspace{3mm}
\vspace{3mm}
\textbf{Context}: \\
$\{\text{context}\}$ \\

Respond with the consolidated and organized list of main ideas and concepts.
\end{tcolorbox}
\caption{The combine prompt in \autoref{fig:savaal-workflow}.}
\label{fig:combine_prompt}
\end{figure}

%%%%%%%%%%%%%%%%%%%%%% Reduce %%%%%%%%%%%%%%%%%%%%%%

\begin{figure}[h]
\centering
\begin{tcolorbox}[width=1\columnwidth, fontupper=\footnotesize,title=Reduce Prompt]
\RaggedRight
\obeylines
\textbf{Instructions}: \\
{You are reducing sets of detailed concept maps, a concise yet comprehensive list of important concepts, generated by extracting concepts from a document and potentially combining subsets of them that are relevant to each other. \\
The goal is to create a structured resource that fully captures the essence of the material for testing and teaching purposes.
\vspace{3mm}
Your task is to:
\begin{itemize}[label=-,nosep]
    \item Identify the most critical concepts from the detailed concept map.
    \item Provide a full-sentence summary for each concept that explains its significance, its relationship to other concepts, and any relevant examples or applications.
    \item Ensure that the summaries are clear, self-contained, and detailed enough to aid in understanding without requiring additional context.
    \item If necessary, combine related concepts into a single summary. Some of the concept maps have broader headings that can be used to guide this process.
\end{itemize}
\vspace{3mm}
\vspace{3mm}
Here is the detailed concept map:} \\
\vspace{3mm}
\vspace{3mm}
\textbf{Context}: \\
$\{\text{context}\}$ \\

Respond with a structured list where each important concept is followed by its full-sentence, detailed summary. For example:
\begin{enumerate}[nosep]
    \item Concept Name: [Detailed full-sentence summary explaining the concept, its relevance, and any examples or applications.]
    \item Another Concept: [Detailed full-sentence summary explaining this concept, its connections to other ideas, and its role in understanding the material.]
\end{enumerate}

Continue in this format for all important concepts.
\end{tcolorbox}
\caption{The reduce prompt in \autoref{fig:savaal-workflow}.}
\label{fig:reduce_prompt}
\end{figure}

%%%%%%%%%%%%%%%%%%%%%% Main Idea Rank %%%%%%%%%%%%%%%%%%%%%%
\begin{figure}[h]
\centering
\begin{tcolorbox}[width=1\columnwidth, fontupper=\footnotesize,title=Ranking Main Ideas]
\RaggedRight
\obeylines
\textbf{Instructions}: \\
{Given the following groups of main ideas extracted from a text, rank them in order of importance, with the most important main idea receiving a rank of 1 and lower ranks for less important ideas. \\
Focus on the most important aspects of the text and the main ideas that are critical to understanding the material. While sometimes important, background information or less critical ideas should be ranked lower. \vspace{3mm}
\textbf{When ranking}:
\begin{itemize}[label=-,nosep]
    \item \textbf{Assign a unique number to each main idea, starting from 1}.
    \item \textbf{Ensure that the most important main idea is ranked first}.
    \item \textbf{Rank the main ideas based on their relevance and significance}.
\end{itemize}}
\vspace{3mm}
Example: \\
\tabto{1cm} Input: [Main Idea 1, Main Idea 2, Main Idea 3] \\
\tabto{1cm} Output: [2, 1, 3] \\
\vspace{3mm}
\vspace{3mm}
\textbf{Main Ideas}: \\
$\{\text{main\_ideas}\}$ \\
\end{tcolorbox}
\caption{The main idea ranking prompt.}
\label{fig:main_idea_ranking_prompt}
\end{figure}

\begin{figure}[h]
\centering
\begin{tcolorbox}[width=1\columnwidth, fontupper=\footnotesize,title=\BaselineMT Additional Question Generation Prompt]
\RaggedRight
\obeylines
\textbf{Instructions}: \\
{
Now, please create $\{\text{num\_questions}\}$ \textbf{additional} multiple-choice questions that require deep understanding, critical thinking, and detailed analysis.  
The questions should go beyond mere factual recall, involving higher-order thinking skills like analysis, synthesis, and evaluation.

\textbf{Provide four answer choices for each question:}
\begin{itemize}[label=-,nosep]
    \item The choices should start with \textbf{A.}, \textbf{B.}, \textbf{C.}, and \textbf{D.}
    \item \textbf{One correct answer.}
    \item \textbf{Three plausible distractors} that are:
    \begin{itemize}[label=$\bullet$,nosep]
        \item Contextually appropriate.
        \item Relevant to the content.
        \item Reflect common misunderstandings or errors without introducing contradictory or irrelevant information.
    \end{itemize}
\end{itemize}
\vspace{3mm}
\textbf{Note:} The questions should focus on one concept and not be overly long.  
\textbf{Note:} The questions should be different from the ones generated in the previous step.
}
\vspace{3mm}
\vspace{3mm}
\textbf{Context}: \\
$\{\text{context}\}$ \\

\end{tcolorbox}
\caption{\BaselineMT Additional Question Generation Prompt.}
\label{fig:baseline_large_question_generation_prompt}
\end{figure}

%%%%%%%%%%%%%%%%%%%%%% Question Structure %%%%%%%%%%%%%%%%%%%%%%

\subsubsection{Evaluation Prompts}
\label{subsubsec:eval-prompts}
The AI evaluation framework consists of six metrics designed to assess multiple-choice questions based on different dimensions. The understanding prompt (\autoref{fig:understanding_prompt}) measures the depth of conceptual understanding required to answer the question. The quality of choices prompt (\autoref{fig:choices_quality_prompt}) evaluates the plausibility of the distractors. The clarity evaluation prompt (\autoref{fig:clarity_prompt}) determines the ambiguity level of the question. The difficulty evaluation prompt (\autoref{fig:difficulty_prompt}) categorizes questions based on their complexity and required cognitive effort. The cognitive level evaluation prompt (\autoref{fig:cognitive_level_prompt}) aligns questions with Bloom's taxonomy~\cite{anderson2001taxonomy}, assessing their level from simple recall to higher-order thinking. Finally, the engagement evaluation prompt (\autoref{fig:engagement_prompt}) measures how stimulating and thought-provoking a question is. Each prompt assigns a score from 1 to 4, ensuring a structured and objective analysis of question quality. We map these numerical scores of 4 to 1 to the qualitative scores of ``Agree'', ``Somewhat Agree'', ``Somewhat Disagree'', and ``Disagree'' for comparison with human evaluation.

%%%%%%%%%%%%%%%%%%%%%% Understanding Prompt %%%%%%%%%%%%%%%%%%%%%%

\begin{figure}[h]
\centering
\begin{tcolorbox}[width=1\columnwidth, fontupper=\footnotesize,title=Understanding Evaluation Prompt]
\RaggedRight
\obeylines

{For the following multiple-choice question:\\
-----------\\
Question: $\{\text{question}\}$\\
\vspace{3mm}
Options: $\{\text{options}\}$\\
\vspace{3mm}
Answer: $\{\text{answer}\}$\\
-----------\\
Please answer the following:\\

Please carefully read the multiple-choice question, the options, and the correct answer.
Rate the understanding level of the question on a scale of 1 to 4 based on the following criteria:
\begin{itemize}[label=-,nosep]
\item \textbf{Score 4} if the question tests a deep understanding of a concept, requiring integration and application of ideas.
\item \textbf{Score 3} if the question tests understanding of a concept but is more straightforward, requiring less integration or application.
\item \textbf{Score 2} if the question largely depends on recall but includes some context-specific details that require a conceptual understanding.
\item \textbf{Score 1} if the question primarily tests memorization of facts or details with minimal to no application of concepts.
\end{itemize}

Please output only a score between 1 and 4.
}

\end{tcolorbox}
\caption{Understanding prompt.}
\label{fig:understanding_prompt}
\end{figure}

%%%%%%%%%%%%%%%%%%%%%% Quality of Choices  Prompt %%%%%%%%%%%%%%%%%%%%%%

\begin{figure}[h]
\centering
\begin{tcolorbox}[width=1\columnwidth, fontupper=\footnotesize,title=Quality of Choices  Evaluation Prompt]
\RaggedRight
\obeylines
{For the following multiple-choice question:\\
-----------\\
Question: $\{\text{question}\}$\\
\vspace{3mm}
Options: $\{\text{options}\}$\\
\vspace{3mm}
Answer: $\{\text{answer}\}$\\
-----------\\
Please answer the following:\\
Please carefully read the multiple-choice question, the options, and the correct answer.  
Rate the quality of choices in the question on a scale of 1 to 4 based on the following criteria:
\begin{itemize}[label=-,nosep]
    \item \textbf{Score 4} if it is challenging to eliminate any incorrect choice due to well-crafted distractors that are plausible, unambiguous, and relevant to the question.
    \item \textbf{Score 3} if incorrect choices can be somewhat challenging to eliminate, requiring a good understanding of the material, but they are less sophisticated.
    \item \textbf{Score 2} if most incorrect choices are fairly easy to eliminate, with perhaps one plausible distractor.
    \item \textbf{Score 1} if incorrect choices are very easy to eliminate, often due to being obviously incorrect or irrelevant.
\end{itemize}

Please output only a score between 1 and 4.
}
\end{tcolorbox}
\caption{Quality of Choices  Evaluation Prompt.}
\label{fig:choices_quality_prompt}
\end{figure}

%%%%%%%%%%%%%%%%%%%%%% Clarity Evaluation Prompt %%%%%%%%%%%%%%%%%%%%%%

\begin{figure}[h]
\centering
\begin{tcolorbox}[width=1\columnwidth, fontupper=\footnotesize,title=Clarity Evaluation Prompt]
\RaggedRight
\obeylines
{For the following multiple-choice question:\\
-----------\\
Question: $\{\text{question}\}$\\
\vspace{3mm}
Options: $\{\text{options}\}$\\
\vspace{3mm}
Answer: $\{\text{answer}\}$\\
-----------\\
Please answer the following:\\

Please carefully read the multiple-choice question, the options, and the correct answer.  
Rate the clarity level of the question on a scale of 1 to 4 based on the following criteria:
\begin{itemize}[label=-,nosep]
    \item \textbf{Score 4} if the question is completely clear and unambiguous.
    \item \textbf{Score 3} if the question is mostly clear, but may have some ambiguity.
    \item \textbf{Score 2} if the question has notable ambiguity that could confuse the reader.
    \item \textbf{Score 1} if the question is highly confusing or unclear.
\end{itemize}

Please output only a score between 1 and 4.
}

\end{tcolorbox}
\caption{Clarity Evaluation Prompt.}
\label{fig:clarity_prompt}
\end{figure}

%%%%%%%%%%%%%%%%%%%%%% Difficulty Evaluation Prompt %%%%%%%%%%%%%%%%%%%%%%

\begin{figure}[h]
\centering
\begin{tcolorbox}[width=1\columnwidth, fontupper=\footnotesize,title=Difficulty Evaluation Prompt]
\RaggedRight
\obeylines

{For the following multiple-choice question:\\
-----------\\
Question: $\{\text{question}\}$\\
\vspace{3mm}
Options: $\{\text{options}\}$\\
\vspace{3mm}
Answer: $\{\text{answer}\}$\\
-----------\\
Please answer the following:\\

Please carefully read the multiple-choice question, the options, and the correct answer.  
Rate the difficulty level of the question on a scale of 1 to 4 based on the following criteria:
\begin{itemize}[label=-,nosep]
    \item \textbf{Score 4} if the question is very challenging, requiring deep understanding and advanced conceptual application.
    \item \textbf{Score 3} if the question is moderately difficult, requiring understanding and some conceptual application.
    \item \textbf{Score 2} if the question is relatively easy and mainly requires recall or basic understanding.
    \item \textbf{Score 1} if the question is very easy and can be answered without specific knowledge.
\end{itemize}

Please output only a score between 1 and 4.
}

\end{tcolorbox}
\caption{Difficulty Evaluation Prompt.}
\label{fig:difficulty_prompt}
\end{figure}

%%%%%%%%%%%%%%%%%%%%%% Cognitive Level Evaluation Prompt %%%%%%%%%%%%%%%%%%%%%%

\begin{figure}[h]
\centering
\begin{tcolorbox}[width=1\columnwidth, fontupper=\footnotesize,title=Cognitive Level Evaluation Prompt]
\RaggedRight
\obeylines

{For the following multiple-choice question:\\
-----------\\
Question: $\{\text{question}\}$\\
\vspace{3mm}
Options: $\{\text{options}\}$\\
\vspace{3mm}
Answer: $\{\text{answer}\}$\\
-----------\\
Please answer the following:\\

Please carefully read the multiple-choice question, the options, and the correct answer.  
Rate the cognitive level of the question based on Bloom's taxonomy on a scale of 1 to 4 based on the following criteria:
\begin{itemize}[label=-,nosep]
    \item \textbf{Score 4} if the question requires higher-level thinking (e.g., analysis, synthesis, or evaluation).
    \item \textbf{Score 3} if the question requires application or understanding of concepts.
    \item \textbf{Score 2} if the question requires basic understanding or recall.
    \item \textbf{Score 1} if the question only tests rote memorization with minimal understanding.
\end{itemize}

Please output only a score between 1 and 4.
}

\end{tcolorbox}
\caption{Cognitive Level Evaluation Prompt.}
\label{fig:cognitive_level_prompt}
\end{figure}

%%%%%%%%%%%%%%%%%%%%%% Engagement Evaluation Prompt %%%%%%%%%%%%%%%%%%%%%%

\begin{figure}[h]
\centering
\begin{tcolorbox}[width=1\columnwidth, fontupper=\footnotesize,title=Engagement Evaluation Prompt]
\RaggedRight
\obeylines
{For the following multiple-choice question:\\
-----------\\
Question: $\{\text{question}\}$\\
\vspace{3mm}
Options: $\{\text{options}\}$\\
\vspace{3mm}
Answer: $\{\text{answer}\}$\\
-----------\\
Please answer the following:\\

Please carefully read the multiple-choice question, the options, and the correct answer.  
Rate the engagement level of the question on a scale from 1 to 4 based on the following criteria:
\begin{itemize}[label=-,nosep]
    \item \textbf{Score 4} if the question is highly engaging and thought-provoking.
    \item \textbf{Score 3} if the question is engaging but not particularly unique or thought-provoking.
    \item \textbf{Score 2} if the question is somewhat engaging but fairly straightforward.
    \item \textbf{Score 1} if the question is uninteresting or not engaging.
\end{itemize}

Please output only a score between 1 and 4.
}

\end{tcolorbox}
\caption{Engagement Evaluation Prompt.}
\label{fig:engagement_prompt}
\end{figure}

%% file: 100refiner.tex
% \clearpage
\subsection{Attempts to Refine Quality of Choices}
\label{appendix:choice-refine}
As shown in human evaluation~\autoref{fig:human-eval-disagree-paper}, the difference between the quality of choice of \Baseline and \TheSystem in short documents is not much. In both systems, the choices are generated alongside the question statement.

To further improve the quality of answer choices, we attempted to use the LLM to refine the incorrect options in the generated questions while keeping the correct answer unchanged, following the prompt in \autoref{fig:refine_prompt}. We evaluated this approach on 100 questions by incorporating the option refiner into \name and conducting a survey with human experts. However, the experts did not favor the refined questions, as the refiner often introduced ambiguity in the incorrect choices or unintentionally made multiple options correct.

%%%%%%%%%%%%%%%%%%%%%% REFINE PROMPT %%%%%%%%%%%%%%%%%%%%%%
\begin{figure}[h]
\centering
\begin{tcolorbox}[width=\columnwidth,fontupper=\footnotesize, title=Option Refinement Prompt]
\RaggedRight
\obeylines
\textbf{Instructions}: \\
{You are given the following information about a multiple-choice question: \\

\textbf{Main Idea}: $\{\text{main\_idea}\}$ \\

\textbf{Relevant Passages}: $\{\text{passages}\}$ \\
\vspace{3mm}
\textbf{Question}: $\{\text{question}\}$ \\

\textbf{Current Options}: $\{\text{options}\}$ \\

\textbf{Correct Answer}: $\{\text{correct\_answer}\}$ \\
\vspace{3mm}
Your task is to refine the three INCORRECT options in a way that:
\begin{itemize}[label=-,nosep]
    \item They remain closely related to the topic of the CORRECT option.
    \item They are incorrect but not obviously off-topic.
    \item They are PLAUSIBLE enough to confuse the reader.
    \item The correct option (and its label) must REMAIN UNCHANGED.
    \item The three incorrect options should ALIGN with the context of the correct answer; 
          for example, if the question asks about advantages, a distractor that lists disadvantages would be considered bad.
\end{itemize}

Return the final question, the NEW options, and the correct answer. \\
\vspace{3mm}
\textbf{REMEMBER}: \\
The correct answer is: $\{\text{correct\_answer}\}$.
}
\end{tcolorbox}
\centering
\caption{The refine prompt used for improving multiple-choice questions.}
\label{fig:refine_prompt}
\end{figure}

%% file: 090examples.tex
\subsubsection{Main Idea Examples}
\label{subsubsec:example_main_idea}

\autoref{subsubsec:example_main_idea} presents examples of the top main ideas extracted from the paper "Attention is All You Need"~\cite{attention_is_all_you_need} in \name (step \circled{3}{customgreen} in \autoref{fig:savaal-workflow}). These main ideas capture some of the key concepts of the paper.

\begin{figure}[t]
\centering
\begin{tcolorbox}[width=1\linewidth, colframe=white, colback=gray!10, sharp corners=southwest, title=]

\begin{tcolorbox}[colframe=black!50, colback=gray!10, title=Transformer model]
The Transformer model is a groundbreaking sequence transduction model that relies entirely on attention mechanisms, eliminating the need for recurrence, and is composed of an encoder-decoder architecture with self-attention and point-wise, fully connected layers, allowing for greater parallelization and efficiency in training.
\end{tcolorbox}

\begin{tcolorbox}[colframe=black!50, colback=gray!10, title=Self-attention mechanism]
Self-attention is a mechanism that relates different positions of a single sequence to compute a representation, and is used in tasks such as reading comprehension, abstractive summarization, and learning task-independent sentence representations.
\end{tcolorbox}

\begin{tcolorbox}
[colframe=black!50, colback=gray!10, title=Positional encoding]
Positional encoding provides information about the order of tokens in a sequence, using fixed sinusoidal functions or learned embeddings, which is crucial for models like the Transformer that lack inherent sequence order awareness.
\end{tcolorbox}

\end{tcolorbox}
\centering
\caption{Main idea examples generated for ``Attention is All You Need''~\cite{attention_is_all_you_need}.}
\label{fig:main_idea_example}
\end{figure}

\subsubsection{Baseline Quiz Example}
\label{subsubsec:example_baseline}
\autoref{fig:baseline_repeat_example} enumerates the questions outputted when prompting an LLM (in this case \gpt) for 20 questions at once. Occasionally, duplicate questions will be output in the same turn. Each pair of duplicated question statements is highlighted in a different color.

\begin{figure*}
\centering
\begin{tcolorbox}[width=\textwidth,fontupper=\footnotesize,title=Repeated Baseline Questions]

\begin{enumerate}
    \item What is the primary advantage of the Transformer model over traditional recurrent neural networks in sequence modeling tasks?
    \begin{itemize}
        \item A. It uses fewer parameters, reducing the model size.
        \item B. It eliminates the need for positional encodings.
        \item C. It allows for more parallelization, improving training efficiency.
        \item D. It requires less data for training, reducing overfitting.
    \end{itemize}
    \textbf{Correct Answer: C. It allows for more parallelization, improving training efficiency.}

    \item How does the Transformer model handle the sequential nature of input data without using recurrence?
    \begin{itemize}
        \item A. By using convolutional layers to capture sequence information.
        \item B. By employing self-attention mechanisms to model dependencies.
        \item C. By using a single attention head to focus on sequence order.
        \item D. By relying on recurrent layers for sequence alignment.
    \end{itemize}
    \textbf{Correct Answer: B. By employing self-attention mechanisms to model dependencies.}

    \item In the context of the Transformer model, what is the purpose of multi-head attention?
    \begin{itemize}
        \item A. To reduce the computational cost of attention mechanisms.
        \item B. To allow the model to attend to different parts of the sequence simultaneously.
        \item C. To increase the dimensionality of the input data.
        \item D. To simplify the model architecture by reducing the number of layers.
    \end{itemize}
    \textbf{Correct Answer: B. To allow the model to attend to different parts of the sequence simultaneously.}

    \item 
        \colorbox{green}{Why is the dot product in the Scaled Dot-Product Attention scaled by \( \frac{1}{\sqrt{d_k}} \)?}
    \begin{itemize}
        \item A. To increase the magnitude of the dot products for better attention.
        \item B. To prevent the softmax function from having extremely small gradients.
        \item C. To ensure the keys and values have the same dimensionality.
        \item D. To make the computation more efficient by reducing the number of operations.
    \end{itemize}
    \textbf{Correct Answer: B. To prevent the softmax function from having extremely small gradients.}

    \item \colorbox{pink}{What role do positional encodings play in the Transformer model?}
    \begin{itemize}
        \item A. They provide a mechanism for the model to learn the order of the sequence.
        \item B. They reduce the dimensionality of the input data.
        \item C. They replace the need for attention mechanisms.
        \item D. They enhance the model's ability to generalize to unseen data.
    \end{itemize}
    \textbf{Correct Answer: A. They provide a mechanism for the model to learn the order of the sequence.}

    \item \colorbox{cyan}{\parbox{0.93\textwidth}{How does the Transformer model ensure that the decoder only attends to previous positions during training?}}
    \begin{itemize}
        \item A. By using a separate attention mechanism for each position.
        \item B. By applying a mask to prevent attention to future positions.
        \item C. By using a recurrent layer to control the flow of information.
        \item D. By training the model on reversed sequences.
    \end{itemize}
    \textbf{Correct Answer: B. By applying a mask to prevent attention to future positions.}

    \item What is the main reason for using residual connections in the Transformer model?
    \begin{itemize}
        \item A. To increase the depth of the model without increasing complexity.
        \item B. To prevent overfitting by adding noise to the model.
        \item C. To facilitate the training of deep networks by mitigating vanishing gradients.
        \item D. To reduce the number of parameters in the model.
    \end{itemize}
    \textbf{Correct Answer: C. To facilitate the training of deep networks by mitigating vanishing gradients.}
\end{enumerate}
\end{tcolorbox}
\end{figure*}

\begin{figure*}\ContinuedFloat
\centering
\begin{tcolorbox}[width=\textwidth,fontupper=\footnotesize,title=Repeated Baseline Questions - Continued]
\begin{enumerate}
    \setcounter{enumi}{7}
    \item Why might self-attention layers be more interpretable than recurrent layers?
    \begin{itemize}
        \item A. They use fewer parameters, making the model simpler to understand.
        \item B. They provide explicit attention distributions that can be analyzed.
        \item C. They rely on fixed positional encodings, reducing variability.
        \item D. They are less sensitive to the order of input data.
    \end{itemize}
    \textbf{Correct Answer: B. They provide explicit attention distributions that can be analyzed.}

    \item In the Transformer model, what is the purpose of using layer normalization?
    \begin{itemize}
        \item A. To increase the model's capacity by adding more layers.
        \item B. To stabilize the training process by normalizing the inputs of each layer.
        \item C. To reduce the model's dependency on positional encodings.
        \item D. To enhance the model's ability to learn long-range dependencies.
    \end{itemize}
    \textbf{Correct Answer: B. To stabilize the training process by normalizing the inputs of each layer.}

    \item \colorbox{yellow}{\parbox{0.93\textwidth}{How does the Transformer model achieve a constant number of operations for relating signals from two arbitrary positions?}}
    \begin{itemize}
        \item A. By using a single attention head for all positions.
        \item B. By employing self-attention mechanisms that connect all positions directly.
        \item C. By using convolutional layers with a large kernel size.
        \item D. By reducing the dimensionality of the input data.
    \end{itemize}
    \textbf{Correct Answer: B. By employing self-attention mechanisms that connect all positions directly.}

    \item What is the primary benefit of using the Adam optimizer in training the Transformer model?
    \begin{itemize}
        \item A. It reduces the need for dropout regularization.
        \item B. It adapts the learning rate based on the training step, improving convergence.
        \item C. It eliminates the need for positional encodings.
        \item D. It simplifies the model architecture by reducing the number of layers.
    \end{itemize}
    \textbf{Correct Answer: B. It adapts the learning rate based on the training step, improving convergence.}

    \item Why is label smoothing used during the training of the Transformer model?
    \begin{itemize}
        \item A. To increase the model's confidence in its predictions.
        \item B. To improve the model's ability to generalize to unseen data.
        \item C. To reduce the model's dependency on positional encodings.
        \item D. To enhance the model's ability to learn long-range dependencies.
    \end{itemize}
    \textbf{Correct Answer: B. To improve the model's ability to generalize to unseen data.}

    \item How does the Transformer model handle the challenge of learning long-range dependencies?
    \begin{itemize}
        \item A. By using a large number of recurrent layers.
        \item B. By employing self-attention mechanisms that connect all positions directly.
        \item C. By using convolutional layers with a large kernel size.
        \item D. By reducing the dimensionality of the input data.
    \end{itemize}
    \textbf{Correct Answer: B. By employing self-attention mechanisms that connect all positions directly.}

    \item What is the main reason for using multiple attention heads in the Transformer model?
    \begin{itemize}
        \item A. To increase the model's capacity by adding more layers.
        \item B. To allow the model to attend to different parts of the sequence simultaneously.
        \item C. To reduce the model's dependency on positional encodings.
        \item D. To enhance the model's ability to learn long-range dependencies.
    \end{itemize}
    \textbf{Correct Answer: B. To allow the model to attend to different parts of the sequence simultaneously.}
\end{enumerate}

\end{tcolorbox}
\end{figure*}

\begin{figure*}\ContinuedFloat
\centering
\begin{tcolorbox}[width=\textwidth,fontupper=\footnotesize,title=Repeated Baseline Questions - Continued]
\begin{enumerate}
    \setcounter{enumi}{14}
    \item Why is the Transformer model considered more efficient than recurrent models for sequence transduction tasks?
    \begin{itemize}
        \item A. It uses fewer parameters, reducing the model size.
        \item B. It eliminates the need for positional encodings.
        \item C. It allows for more parallelization, improving training efficiency.
        \item D. It requires less data for training, reducing overfitting.
    \end{itemize}
    \textbf{Correct Answer: C. It allows for more parallelization, improving training efficiency.}
    
    \item \colorbox{cyan}{\parbox{0.93\textwidth}{How does the Transformer model ensure that the decoder only attends to previous positions during training?}}
    \begin{itemize}
        \item A. By using a separate attention mechanism for each position.
        \item B. By applying a mask to prevent attention to future positions.
        \item C. By using a recurrent layer to control the flow of information.
        \item D. By training the model on reversed sequences.
    \end{itemize}
    \textbf{Correct Answer: B. By applying a mask to prevent attention to future positions.}

    \item What is the primary advantage of using self-attention over convolutional layers in the Transformer model?
    \begin{itemize}
        \item A. Self-attention requires fewer parameters, reducing model size.
        \item B. Self-attention allows for more parallelization, improving training efficiency.
        \item C. Self-attention captures long-range dependencies more effectively.
        \item D. Self-attention eliminates the need for positional encodings.
    \end{itemize}
    \textbf{Correct Answer: C. Self-attention captures long-range dependencies more effectively.}

    \item \colorbox{green}{Why is the dot product in the Scaled Dot-Product Attention scaled by \( \frac{1}{\sqrt{d_k}} \)?}
    \begin{itemize}
        \item A. To increase the magnitude of the dot products for better attention.
        \item B. To prevent the softmax function from having extremely small gradients.
        \item C. To ensure the keys and values have the same dimensionality.
        \item D. To make the computation more efficient by reducing the number of operations.
    \end{itemize}
    \textbf{Correct Answer: B. To prevent the softmax function from having extremely small gradients.}

    \item \colorbox{pink}{What role do positional encodings play in the Transformer model?}
    \begin{itemize}
        \item A. They provide a mechanism for the model to learn the order of the sequence.
        \item B. They reduce the dimensionality of the input data.
        \item C. They replace the need for attention mechanisms.
        \item D. They enhance the model's ability to generalize to unseen data.
    \end{itemize}
    \textbf{Correct Answer: A. They provide a mechanism for the model to learn the order of the sequence.}

    \item \colorbox{yellow}{\parbox{0.93\textwidth}{How does the Transformer model achieve a constant number of operations for relating signals from two arbitrary positions?}}
    \begin{itemize}
        \item A. By using a single attention head for all positions.
        \item B. By employing self-attention mechanisms that connect all positions directly.
        \item C. By using convolutional layers with a large kernel size.
        \item D. By reducing the dimensionality of the input data.
    \end{itemize}
    \textbf{Correct Answer: B. By employing self-attention mechanisms that connect all positions directly.}
\end{enumerate}
\end{tcolorbox}
\caption{An example of repeated questions using the baseline method. Duplicated questions are highlighted in the same color.}
\label{fig:baseline_repeat_example}
\end{figure*}